\documentclass[pdflatex,sn-mathphys-ay]{sn-jnl}% Math and Physical Sciences Author Year Reference Style
\usepackage{graphicx}%
\usepackage{multirow}%
\usepackage{amsmath,amssymb,amsfonts}%
\usepackage{amsthm}%
\usepackage{mathrsfs}%
\usepackage[title]{appendix}%
\usepackage{xcolor}%
\usepackage{textcomp}%
\usepackage{manyfoot}%
\usepackage{booktabs}%
\usepackage{algorithm}%
\usepackage{algorithmicx}%
\usepackage{algpseudocode}%
\usepackage{listings}%
\usepackage{hyperref}
\usepackage{subcaption}
\begin{document}

\title[Detecting Bias in Breast Cancer Detection AI]{
%A comprehensive analysis of the EMBED dataset and its biases using a novel AI model
%Impact of Demographic Biases on the Performance of a Breast Cancer Detection: A Study Using the EMBED Dataset
Detecting and Monitoring Bias for Subgroups in Breast Cancer Detection AI
%Detecting and Monitoring Bias for Subgroups in Breast Cancer Detection AI: A Study using the EMBED and RSNA Mammography datasets
%Monitoring for clinical AI bias: Method to detect subgroup performance drifts in breast cancer detection 
}

% Detecting Bias for subgroups in BCD AI models: A study using the EMBED and RSNA mammography datasets


\author*[1,5]{\fnm{Amit Kumar} \sur{Kundu}}\email{amit314@umd.edu}
%\author[2]{\fnm{Ghada} \sur{Zamzmi}}\email{alzamzmigaa@fda.hhs.gov}
\author[3]{\fnm{Florence X.} \sur{Doo}}\email{FDoo@som.umaryland.edu}
\author[4,5]{\fnm{Vaishnavi} \sur{Patil}}\email{vspatil@umd.edu}
%\author[2]{\fnm{Aldo} \sur{Badano}}\email{aldo.badano@fda.hhs.gov}
\author[4,5,6]{\fnm{Amitabh} \sur{Varshney}}\email{varshney@umd.edu}
\author[1,5,6]{\fnm{Joseph} \sur{Jaja}}\email{josephj@umd.edu}

\affil*[1]{\orgdiv{Department of Electrical and Computer Engineering}, \orgname{University of Maryland}%, \orgaddress{\street{Street}, \city{City}, \postcode{100190}, \state{State}, \country{Country}}
}

%\affil[2]{\orgdiv{Office of Science and Engineering Laboratories, Center for Devices and Radiological Health}, \orgname{U.S. Food and Drug Administration}%, \orgaddress{\street{Street}, \city{City}, \postcode{610101}, \state{State}, \country{Country}}
%}

\affil[3]{\orgdiv{Department of Diagnostic Radiology and Nuclear Medicine}, \orgname{University of Maryland School of Medicine}%, \orgaddress{\street{Street}, \city{City}, \postcode{610101}, \state{State}, \country{Country}}
}

\affil[4]{\orgdiv{Department of Computer Science}, \orgname{University of Maryland}%, \orgaddress{\street{Street}, \city{City}, \postcode{610101}, \state{State}, \country{Country}}
}

\affil[5]{\orgdiv{Institute for Health Computing}, \orgname{University of Maryland}%, \orgaddress{\street{Street}, \city{City}, \postcode{610101}, \state{State}, \country{Country}}
}

\affil[6]{\orgdiv{University of Maryland Institute for Advanced Computer Studies}%, \orgname{University of Maryland}, \orgaddress{\street{Street}, \city{City}, \postcode{610101}, \state{State}, \country{Country}}
}


\abstract{
Automated mammography screening plays an important role in early breast cancer detection. However, current machine learning models, developed on some training datasets, may exhibit performance degradation and bias when deployed in real-world settings. 
In this paper, we analyze the performance of high-performing AI models on two mammography datasets—the Emory Breast Imaging Dataset (EMBED) and the RSNA 2022 challenge dataset. Specifically, we evaluate how these models perform across different subgroups, defined by six attributes, to detect potential biases using a range of classification metrics. Our analysis identifies certain subgroups that demonstrate notable underperformance, highlighting the need for ongoing monitoring of these subgroups' performance. To address this, we adopt a monitoring method designed to detect performance drifts over time. Upon identifying a drift, this method issues an alert, which can enable timely interventions. This approach not only provides a tool for tracking the performance but also helps ensure that AI models continue to perform effectively across diverse populations.  }

%%================================%%
%% Sample for structured abstract %%
%%================================%%




\maketitle

\section{Introduction}\label{sec:intro}
Early breast cancer detection (BCD) through mammography screening continues to be a major focus in radiology as it plays a critical role in reducing mortality rates (\cite{coleman2017early, ginsburg2020breast}).
%However, manually screening mammograms by practitioners makes it expensive and potentially prone to human errors (\cite{evans2013if,maxwell1999breast}). Moreover, shortage of skilled radiologists can exacerbate the issue (\cite{konstantinidis2024shortage, wing2009workforce}). 
Although artificial intelligence (AI) models can help radiologists to evaluate mammograms (\cite{sahu2023recent,evans2013if,maxwell1999breast}), training such models face the challenge of limited datasets that may not fully represent all subgroups or cover variations in data distributions.
%and that are subject over time to possible data drifts and changes in scanning technologies and acquisition protocols.
%which can potentially lead to an enhanced diagnostic accuracy and reduced human error (). 
%The Radiological Society of North America (RSNA) and other sources host mammogram datasets aiming to improve the performance BCD models (\cite{rsna-breast-cancer-detection, halling2020optimam, pham2022vindr, lekamlage2020mini, heath1998current}). 
%training such models face the challenge of limited datasets 
%that are restricted in scope and demographic diversity and that 
%that cannot reflect future data distributions and drifts.%and changes. 

Historically, certain racial groups face barriers to healthcare access because of many socio-economic factors (\cite{azin2023racial, hershman2005racial, hussain2004ethnic}). This lack of access can result in datasets that do not adequately represent these groups, potentially cause AI models to show biases for these groups. Even with seemingly balanced datasets, subtle biases may persist in the collected data due to systemic inequalities in the quality of healthcare (\cite{obermeyer2019dissecting}). Among these groups, African American patients are often underrepresented in both breast imaging and broader healthcare datasets (\cite{yedjou2019health,newman2017health}). Moreover, studies show that models trained with well optimized design choices on a limited dataset performs well on data similar to the training distribution but struggle when tested on different demographics or varying imaging characteristics (\cite{de2019does,wilson2019predictive}). 
This raises concerns about the performance of such AI models across different demographic and other sensitive attributes in real-world clinical practice.

Several studies investigated the impact of potential biases on the performance of AI models across different demographic and clinical subgroups (\cite{sun2022performance, afzal2023towards, mehta2024evaluating}). 
Addressing these biases through a comprehensive analysis of these models can help developers build more reliable systems that can be refined over time to improve healthcare outcomes for all patients.
%By comprehensively analyzing the AI models and by addressing the biases, developers can build reliable AI models that can be refined as needed to improve healthcare outcomes for all patients. 
%Unfortunately, BCD from mammography lacks such reliable AI models. 
The development of such a framework necessitates large datasets that cover diverse demographics with relevant clinical information. Recently, \cite{nguyen2024patient} analyzed the performance of a commercially available model exclusively on negative Digital Breast Tomosynthesis (DBT) examinations across different demographic groups. While the study highlights discrepancies in false positive rates among certain subgroups, it does not address the characteristics of the model's training data, leaving questions about potential sources of bias unanswered.
%Recently, \cite{nguyen2024patient} analyzes the performance of a commercially available model only on the negative Digital Breast Tomosynthesis (DBT) examinations for different demographic groups. Although the study indicates discrepancy in false positive rates towards some subgroups, there is no discussion related to the model's training data characteristics.

Moreover, during clinical deployment, BCD-AI models are likely to encounter data distributions that differ from their training data due to shifts in subgroup composition, updates in scanning technology, and site-specific variations in imaging protocols and clinical workflows. This may result in a reduced or biased model performance, which can lead to missed diagnoses or delayed treatments. Therefore, an appropriate performance monitoring method is of high significance to quickly detect performance degradations and to ensure clinically actionable insights necessary for follow up adjustments.

This paper investigates the reliability of high-performing AI models in BCD by analyzing their performance across different demographic and clinical subgroups. To achieve this, we leverage two large and diverse datasets—the RSNA Screening Mammography Breast Cancer Detection dataset (\cite{rsna-breast-cancer-detection}) and the EMory BrEast Imaging (EMBED) dataset (\cite{jeong2023emory})—which contain extensive patient information. This comprehensive analysis provides a unique opportunity to assess potential biases in AI models and gain deeper insights into the characteristics of these datasets (\cite{sweeney2022machine}).

%This paper leverages the RSNA Screening Mammography Breast Cancer Detection (\cite{rsna-breast-cancer-detection}) and the EMory BrEast Imaging (EMBED) (\cite{jeong2023emory}) Datasets, large and diverse collections of mammograms with extensive patient information to conduct a thorough analysis of state of the art AI models on the performance related to different groups. This provides a unique opportunity to understand these datasets, to investigate the reliability of AI models for digital mammography by identifying potential biases across a wide range of demographic and other relevant subgroups (\cite{sweeney2022machine}). 

\begin{figure}[!t]
\centering
\includegraphics[width=0.8\linewidth]{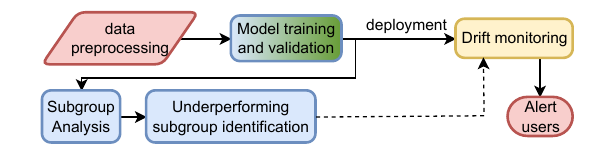}
\caption{Overview of the proposed framework. 
%The framework focuses on assessing BCD model performance using a diverse dataset, and outlines how subgroup analysis can be performed to understand the impacts of different attributes before deploying the models, which can be monitored to detect bias using methods such as CUSUM and send alert.
The framework assesses BCD performance across subgroups to understand the impacts of different attributes, enabling pre-deployment bias identification and post-deployment drift monitoring for alerts.}\label{fig:BLOCK}
\end{figure}

We begin by analyzing the performance of high-performing AI models on the RSNA and EMBED datasets, focusing on identifying and evaluating performance across demographic and clinical subgroups. To explore these biases, we develop our own EMBED-AI model, which achieves strong results across multiple performance metrics. Despite its overall high performance, we demonstrate that biases in the dataset can lead to disparities in clinical outcomes for certain subgroups. To address this, we implement a performance monitoring method based on statistical process control (SPC) from \cite{zamzmi2024out, prathapan2024quantifying, feng2022clinical}, enabling the detection of model performance degradations over time. Fig. \ref{fig:BLOCK} provides an overview of the proposed framework. By systematically identifying disparities through subgroup analysis and tracking their performance over time, this framework allows for ongoing refinement of the model to ensure both reliability and fairness for all patients.

%We start by describing high-performing BCD models on the two datasets with a focus on identifying and evaluating performance across demographic and clinical subgroups. In particular, we develop our own  EMBED-AI model that achieves very good results across different performance metrics.  We demonstrate that biases in the dataset could lead to potential disparities in clinical outcomes for some subgroups despite the models' overall high performance.  Then, we adopt a performance monitoring method based on statistical process control (SPC) from \cite{zamzmi2024out, prathapan2024quantifying,feng2022clinical}, to detect degradations in model performance over time. 

The main contributions of the paper can be summarized as follows:
\begin{itemize}
\item We train a new AI model on the large EMBED dataset using advanced computer vision techniques, such as strong data augmentations, cosine annealing, label smoothing, dropout with drop-path regularization, the exponential moving averaging of model weights and training with a significant model architecture while optimizing performance within limited computational resources. The model achieves excellent results across multiple performance metrics, including Positive Predictive Value (PPV), sensitivity, F1-score, and Area Under the Receiver Operating Characteristic Curve (AUROC).
%We develop a new AI model on the large EMBED dataset using advanced computer vision technologies, optimizing its performance using limited computational resources. The new model demonstrates significant results across different performance metrics, such as PPV, sensitivity, F1-score, and AUROC.

%We adopt a state-of-the-art BCD-AI model on the RSNA dataset and leverage transfer learning to develop a new AI model on the relatively new EMBED dataset, optimizing performance on limited computational resources. The two models, especially the new EMBED AI model, demonstrate significant results across different performance metrics. 
%We employ transfer learning techniques to train a BCD-AI model on the large EMBED dataset using limited computational resources. The model achieves remarkable results using a wide range of performance metrics.

\item We conduct a rigorous performance analysis across subgroups in the RSNA dataset using its top-performing model and in the EMBED dataset using our trained model. The subgroups are defined based on key attributes, including age, race, breast tissue density, mammogram scanning technology, view position, and clinical site. Through this analysis, we identify subgroups for which each model exhibits bias and discuss their clinical implications. 

%\item We conduct a rigorous performance analysis across subgroups from the RSNA dataset using its top-performing model and the EMBED dataset using our trained model. We consider the subgroups based on a number of important attributes, such as participant age, race, breast tissue density, mammogram scanning technology, view position, and site.
%We rigorously analyze the performance on different subgroups from both the RSNA and the EMBED datasets based on a number of important attributes, such as participant age, race, breast tissue density, mammogram model, view position and site. 
%As a result, we identify subgroups towards which each of the models is biased and discuss their clinical implications.
%Given the observed variability in the dataset and the subgroup performance, EMBED appears to be more suitable for training BCD-AI models.

%\item We analyze the statistical associations between each considered attribute with the targets and the predicted labels. Significant associations are found only between age groups and the targets for both datasets. Furthermore, the associations between attributes and predicted labels closely mirror those between attributes and target indicating that the models are mostly capturing the interrelationships within the datasets. 

%Therefore, the observed performance disparities happen primarily due to the bias inherited from the training data. 

\item We adopt Statistical Process Control (SPC) methods, such as the cumulative sum (CUSUM) control chart, as a performance monitoring approach to enable continuous tracking and detection of performance drifts. This method can enable practitioners to systematically assess model stability over time. Our findings demonstrate that CUSUM can effectively and rapidly identify performance drifts in underperforming subgroups.

%\item We adopt a performance monitoring method for the BCD-AI models. Such a method can help practitioners to continuously monitor and detect drifts in model performance. We find that SPC methods such as cumulative sum (CUSUM) control chart can effectively and quickly identify performance drifts for underperforming subgroups.

\end{itemize}

The remainder of the paper is structured as follows: Section \ref{material} introduces the datasets, evaluation protocols, and experimental setup. In Section \ref{results}, we present the overall performance of each model, followed by the results of the subgroup analysis and an evaluation of the performance monitoring method. Finally, Section \ref{conclusion} provides concluding remarks.

\section{Materials and Methods}\label{material}
In this section, we present the datasets, the corresponding AI models, and the evaluation metrics used in our analysis. We also describe the method for monitoring performance.

\subsection{Detection Task}
We frame the AI task as a binary classification problem from a mammogram. The model for the RSNA dataset (\cite{rsna-breast-cancer-detection}) is trained on the defined \textit{Cancer} variable as target. For EMBED, we define the target using the well-established Breast Imaging Reporting and Data System (BIRADS) scores of the mammograms (\cite{sickles2013acr}).
Physicians assign BIRADS scores (from 0 to 6) to the mammograms in a standard way as described in Table \ref{BIRADS_definition} in Appendix \ref{birads_def}.
We combine the benign and normal findings (BIRADS scores of 1 and 2) to construct the \textit{Negative} class, and the suspicious through the malignancy findings (BIRADS scores of 4, 5 and 6) to create the \textit{Positive} class. We categorize mammograms labeled as suspicious or highly suspicious under the \textit{Positive} class, as these cases typically require further evaluation through biopsy to determine the final diagnosis. Mammograms with BIRADS 0 are excluded from the classification task, as these mammograms necessitate follow-up diagnostic imaging; however, we include the follow up diagnostic images in the dataset. %While such a classification target is not ideal, we see no viable alternative given the information provided by the EMBED dataset.

\subsection{Datasets}
%The datasets commonly used to train BCD-AI models in the literature have several limitations. They often include an insufficient number of mammograms, particularly with few positive cases (\cite{khaled2021categorized, cui2021chinese, pham2022vindr, lekamlage2020mini, heath1998current}), which make them unsuitable for our comprehensive analysis. 

The datasets commonly used to train BCD-AI models often lack sufficient mammograms, especially positive cases (\cite{khaled2021categorized, cui2021chinese, pham2022vindr, lekamlage2020mini, heath1998current}), making them unsuitable for comprehensive analysis.

To address this limitation, the Radiological Society of North America (RSNA) hosted in 2022 a challenge using a dataset of 54,706 screening mammograms with a prevalence\footnote{Prevalence is defined as the ratio of positive samples to the total number of samples in a dataset.} of 2.12\% to facilitate the training of BCD-AI models (\cite{rsna-breast-cancer-detection}). While the dataset includes important attributes such as tissue density, view position, and age, it lacks diagnostic mammograms and does not provide racial or ethnic descriptions of the participants. The details of the dataset for different attributes are outlined in Table \ref{tab:num_samples}. The absence of racial information may limit the dataset's representation across all racial groups, potentially excluding those with a higher prevalence of breast cancer.

Recently, Emory University School of Medicine released the EMBED mammography dataset, which is large, granular and covers racially diverse demographics (\cite{jeong2023emory}). The publicly available portion contains a large number of screening and diagnostic mammograms from a large number of patients. The dataset provides information related to different relevant attributes with pathological reports where appropriate. We only consider 2D mammograms with MLO and CC view positions resulting in a set of 217672 mammograms from 20128 participants and 56480 examinations. Mammogram examples are presented in Fig. \ref{fig:BIRADS_samples} and the distributions of these attributes are provided in Table \ref{tab:num_samples}. 
We discuss the attribute distributions along with the performance analysis in details in Section \ref{sub_results}. 
Moreover, to understand the inter-relationship between the considered attributes within dataset, the statistical associations between attributes for both datasets are presented later in Table \ref{tab:data_corr}.

%We construct the set of \textit{normal} examples from the screening mammograms and the set of \textit{cancer} examples from the diagnostic ones, ensuring a manageable dataset size. 
%We drop the duplicate mammograms from the collected sets with inconsistent BIRADS scores. 

Table \ref{tab:num_samples} indicates that both EMBED and RSNA are highly imbalanced datasets with a prevalence of 3.37\% and 2.12\%, respectively. Moreover, considering the number of samples and prevalence of subgroups, the RSNA dataset distribution differs significantly from EMBED. 
%,  we evaluate the F1 score of the entire training set for each fold in the range of  to find the best threshold with the highest F1 score. This threshold is fixed for the test set for a fair comparison throughout the entire analysis.

% We train our model on EMBED and test it on the subgroups extracted based on races, age groups, breast tissue densities, and mammogram views.

\begin{table*}[!t]\centering
\scriptsize
\caption{Distribution of BIRADS scores and attributes in the filtered EMBED and the RSNA Datasets.}
   \begin{tabular}{l|c|c|c||c|c|c}
    \hline
&\multicolumn{3}{c||}{EMBED} &\multicolumn{3}{c}{RSNA} \\\hline
&(+) ves &(-) ves &preval- &(+) ves &(-) ves &preval- \\
&&&ence (\%)&& &ence (\%) \\\hline
Whole set &7346 &210326 &3.37 &1158 &53548 &2.12 \\ \hline
BIRADS Score & \multicolumn{6}{c}{} \\ \hline
0 & & & &664 &7585 &8.05 \\
1 & &192351 & & &15772 &-- \\
2 & &17975 & & &2265 &-- \\
4 &6263 & & & & & \\
5 &368 & & & & & \\
6 &715 & & & & & \\ \hline
Subgroup & \multicolumn{6}{c}{} \\ \hline
Age Group & \multicolumn{6}{c}{} \\ \hline
age $<40$ &936 &2390 &28.14 &5 &471 &1.05 \\
$40 \le \text{age} < 50$ &1848 &44428 &3.99 &90 &9958 &0.90 \\
$50 \le \text{age} < 60$ &1693 &57850 &2.84 &281 &18383 &1.51 \\
$60 \le \text{age} < 70$ &1519 &59687 &2.48 &418 &17585 &2.32 \\
$40 \le \text{age} < 80$ &874 &36483 &2.34 &306 &6042 &4.82 \\
age $\ge 80$ &476 &9488 &4.78 &58 &1109 &4.97 \\ \hline
Race & \multicolumn{3}{c||}{} & \multicolumn{3}{l}{Site} \\ \hline
Caucasian &2822 &92462 &2.96 & \begin{tabular}{c|c} 1 & 664\end{tabular} &28855 &2.25 \\
AAB &3693 &92390 &3.84 &\begin{tabular}{c|c} 2 & 494 \end{tabular} &24693 &1.96 \\ 
Asian &345 &11163 &3.00 & & & \\
AIAN &19 &365 &4.95 & & & \\
NHPI &67 &1604 &4.01 & & & \\
Multiple &14 &551 &2.48 & & & \\
Unknown &386 &11791 &3.17 & & & \\ \hline
Tissue density & \multicolumn{6}{c}{} \\ \hline
1 &558 &26335 &2.07 &53 &3052 &1.71 \\
2 &2460 &91133 &2.63 &309 &12342 &2.44 \\
3 &3521 &82025 &4.12 &277 &11898 &2.28 \\
4 &598 &10307 &5.48 &25 &1514 &1.62 \\ \hline
View position & \multicolumn{6}{c}{} \\ \hline
MLO &3145 &112528 &2.72 &590 &27313 &2.11 \\
CC &4201 &97798 &4.12 &566 &26199 &2.11 \\ \hline
Scanner & \multicolumn{3}{c||}{} & \multicolumn{3}{l}{Machine ID for RSNA} \\ \hline
Lorad Selenia &176 &7197 &2.39 & \begin{tabular}{l|c} 021 & 154\end{tabular} &8067 &1.87 \\
Selenia Dimensions &6680 &185708 &3.47 &\begin{tabular}{l|c} 029 & 161\end{tabular} &8106 &1.95 \\
S2000D ADS17.4.5 &237 &7634 &3.01 &\begin{tabular}{l|c} 048 & 179\end{tabular} &8520 &2.06 \\
S2000D ADS17.5 &34 &1094 &3.01 &\begin{tabular}{l|c} 049 & 614\end{tabular} &22915 &2.61 \\
S Ess. ADS53.40 &197 &8313 &2.31 &\begin{tabular}{l|c} 093 & 14\end{tabular} &1901 &0.73 \\
S Pristina &19 &380 &4.76 &\begin{tabular}{l|c} 170 & 23\end{tabular} &900 &2.49 \\
(S for Senograph) & & & &\begin{tabular}{l|c} 190 & 05\end{tabular} &140 &3.45 \\
\hline
\end{tabular}\label{tab:num_samples}
\end{table*}

\subsection{Training Method}\label{Model_training}
We focus on the top-performing BCD-AI model developed by \cite{dangnh0611} for the RSNA 2022 challenge (\cite{rsna-breast-cancer-detection}). This model is built on a lightweight ConvNeXt backbone architecture (\cite{liu2022convnet}) and is trained on both the challenge dataset and additional external mammography datasets, such as VinDr-Mammo, MiniDDSM, CMMD, BMCD and CDD-CESM datasets (see Section \ref{related_work}). ConvNeXt, a high-capacity convolutional neural network, is designed to offer competitive performance with Vision Transformers and Swin Transformers while maintaining lower computational complexity, making it well-suited for large-scale computer vision tasks.

%However, as demonstrated later, the model by \cite{dangnh0611} underperforms on the EMBED dataset for the defined task.

For EMBED dataset, we train our model using strong data augmentations (Appendix \ref{sec:aug}), cosine annealed learning rate scheduling (\cite{loshchilov2016sgdr}), label smoothing for positive samples (\cite{lukasik2020does}), dropout with drop-path regularization (\cite{huang2016deep}), the exponential moving averaging of model weights (\cite{morales2024exponential}) and training with the ConvNext model (\cite{liu2022convnet}). To efficiently train the model within the available computational resources (four A5000 GPUs), we use the pretrained model by \cite{dangnh0611} and apply transfer learning to achieve a significant performance improvement. In particular, we freeze the lower layers of the pretrained model, which are effective at learning low-level features relevant to mammography, allowing them to serve as rich mammogram feature extractors. We then train only the final stage and the head of the model, resulting in 15 million trainable parameters out of a total of 50 million. Training the model on such a diverse dataset with advanced training techniques  makes our model suitable for the comprehensive analysis. The model outputs a probability of a positive prediction through a single output node with sigmoid activation. 
%An exponential moving average (EMA) model is tracked and used for the final evaluation. 
To address data imbalance, we employ a balanced batch sampler, maintaining a 1:1 ratio of positive to negative samples. Further augmentation, training and preprocessing details of each mammogram to extract $1024 \times 2048$ region of interest (ROI) input are described in Appendix \ref{app:train_details}. 
%Both the pretrained model by \cite{dangnh0611} and our refined one are trained in a similar way with good training strategies and design choices (discussed in Appendix \ref{app:train_details}), which makes them suitable for our comprehensive analysis. 
%The details of training strategies, choice of hyperparameters and the set augmentations are discussed in Appendix \ref{app:train_details}.
A four-fold cross validation is performed. We observe similar BIRADS distributions and subgroup distributions for each attribute across the folds even though we do not consider any attribute information for data splitting. 

% mention techniques for EMBED with citation  and a concluding sentence

\subsection{CUSUM-based Performance Monitoring}
CUSUM control chart is a powerful tool to continuously monitor and control the quality of a statistical process (\cite{crosier1986new}).
This tool can be used to drift in data characteristics or detect degradation in model performance over time as demonstrated in \cite{prathapan2024quantifying, zamzmi2024out}. In this work, we employ a CUSUM-based method to monitor performance drift in under-represented or underperforming subgroups. This approach helps flags potential biases in model performance and prevent unfair evaluations.

We track drift in a specific performance metric, such as sensitivity of the model, which is assessed at the batch level. Initially, our model processes batches that closely resemble the data distribution seen during training. Over time, it encounters shifted distributions that may lead to performance degradation. To evaluate our monitoring method, we create a simulated scenario with $N$ batches as a sequential data. The batches comprise of $B$ mammograms with bootstrapped sampling. 
For simulating the monitoring of performance degradation due to changes in data distribution of one attribute (say, age), we first empirically compute the base marginal distribution $p_{base} = [p_0, p_1, ..., p_l]$ for the attribute from the overall training data, where $p_l$ is the proportion of $l$-th subgroup in the training data.

We assume that the first $N/2$ batches are constructed from $p_{base}$ with a small flexibility (say, $\pm 2.5\%$) around the base proportions. 
Starting halfway through the process, we start introducing deviations in attribute distribution by removing or adding more samples from a certain subgroup $u$ to each batch at a portion $p'_u = p_u + \Delta p$. $u$ is selected from the subgroups and $\Delta p$ is varied within a range. This simulation is designed to detect when the metric scores are below the acceptable range with CUSUM charts by varying the deviations of data distribution in amount and direction.

The CUSUM control signals $S_i^{u}$ and $S_i^{l}$, to detect an upward or a downward shift, are computed as:
\begin{align}
S_i^{u} =\text{max}(0,S_{i-1}^{u} +(m_{i} - \mu -k)) \notag\\
S_i^{l} =\text{min}(0,S_{i-1}^{l} +(m_{i} - \mu +k)) \notag
\end{align}

where $m_i$ is the computed metric for $i$-th batch, $\mu$ is the mean value of the metric, and $k$ is an allowance. 
%$S_i^{u}$ and $S_i^{l}$ are the control signals to detect an upward or a downward shift. 
A drift is detected when one of the control signals is outside the threshold range $(-h,h)$. $h=4\sigma$, where $\sigma$ is the standard deviation of the metric. Prior to deployment, we only need to calculate the mean $\mu$ and standard deviation $\sigma$, which are derived from the training batches. The sensitivity of the CUSUM control chart is fine-tuned by adjusting the parameters $k$ and $h$, which strikes a balance between false alarm and delays in drift detection.

\subsection{Evaluation Metrics}
We used three set of metrics: classification metrics, statistical significance metrics, and monitoring metrics as described next. 

%We present the classification metrics employed to evaluate the two models, along with the statistical significance tests used to analyze the models' performance across the attributes.

%\subsubsection{Classification Metrics} 
The performance of the models are evaluated using positive predictive value (PPV or precision), sensitivity (or recall), F1 score, accuracy, specificity, area under the receiver operating characteristic curve (AUROC), and Monte Carlo (MC) dropout uncertainty. 
Sensitivity is important due to the low true positive detection rates in breast cancer diagnosis and PPV is important because of low success rates in positive predictions.
To select the best detection threshold, we search for the highest training F1 score within the threshold range of $\{0.01, 0.02, ..., 1.0\}$ and compute all metrics using the fixed threshold during testing. Selecting the threshold based on the highest F1 score creates a balance between PPV and sensitivity for such an imbalanced dataset scenario.
%Precision represents the proportion of correct examples among the detected positives, while recall measures the true positive detection rate. 

%\subsubsection{Statistical Significance Tests}

%We perform the Wilcoxon rank-sum test to evaluate differences between the cancer prediction probability and the cancer positivity.

For attributes divided into exactly two subgroups, we use the Mann-Whitney U test (\cite{mann1947test}) to measure the statistical significance of differences in performance between these subgroups.. For attributes categorized into more than two subgroups, we use the Kruskal-Wallis test (\cite{kruskal1952use}) to determine if significant differences exist among the groups. If significant, we then conduct Dunn's posthoc analysis (\cite{dunn1964multiple}) for detailed pairwise comparisons.%For attributes with more than two subgroups, we apply the Kruskal-Wallis test (\cite{kruskal1952use}), followed by Dunn's posthoc analysis (\cite{dunn1964multiple}) for pairwise comparisons. 
To assess correlations between two continuous variables, we apply Pearson's correlation test and to assess associations between two categorical variables, we perform the Cramer's V association test. The significance level is set at 0.05 for all analyses.

We assessed the performance of CUSUM in detecting drift using two key metrics: the False Alarm Rate (FAR) and Detection Delay. FAR measures how frequently the method incorrectly signals a change when there is no actual drift. A low FAR is desirable as it indicates that the method minimizes unnecessary alerts, thus avoiding the costs and disruptions of false positives. Detection Delay, on the other hand, measures the time or the number of data points required for the method to identify a true change once it has occurred. Minimizing detection delay ensures that any significant changes are recognized promptly, allowing for quick response to actual drift.

%\subsubsection{Fairness Metrics}
%We used fairness metrics to summarize potential biases across subgroups using disparate impact, equal opportunity, and predictive parity (\cite{verma2018fairness,DBLP:journals/corr/FriedlerSV16}). \textit{Disparate impact} is calculated as the ratio of the minimum positive selection rate to the maximum positive selection rate across subgroups. \textit{Equal opportunity} measures the consistency of true positive rates across subgroups and is calculated as the difference between the maximum and minimum sensitivity. \textit{Predictive parity} ensures that the likelihood of a positive prediction being correct is consistent across subgroups, measured as the ratio of the minimum PPV to the maximum PPV.

\begin{table*}[t]\centering
\caption{Performance [average(standard deviation)] on the overall datasets for their respective targets}\label{tab:overall_results}
\scriptsize
\begin{tabular}{lccccc}\hline
Model &dataset&PPV &sensitivity & F1 score & AUROC  \\\hline
\cite{dangnh0611} &RSNA&0.548(0.099) &0.500(0.07) &0.523(0.08) & 0.916(0.035)  \\
%\cite{dangnh0611} &EMBED&0.2821 &0.2543 &0.2670 & 0.7582 \\
This work &EMBED&0.896(0.01) &0.679(0.014) &0.773(0.006) & 0.945(0.001) \\
\hline
\end{tabular}
\end{table*} 

\section{Results and Discussions}\label{results}
We present three sets of results. The first set shows the performance of the models on the overall datasets. We then show the results of analyzing the models' consistency across subgroups. Finally, we demonstrate the effectiveness of applying CUSUM to detect drift in the models' performance.

\subsection{Model Performance}
Before conducting subgroup analysis, it is important to ensure the models achieve acceptable performance on the respective datasets. Therefore, we evaluate the performance of the model developed by \cite{dangnh0611} on the RSNA dataset and our trained model (described in Section \ref{Model_training}) on the EMBED dataset, with the results presented in Table \ref{tab:overall_results}. We observe that the model by \cite{dangnh0611} achieves an AUROC of 91.6\% on RSNA.
%Using our trained model as described in Section \ref{Model_training} we obtain the results on EMBED shown in Table \ref{tab:overall_results}. 
The Table also shows that our model achieves good performance on EMBED across different metrics. Moreover, our model outputs significantly higher probability scores for the positive cases than the negatives cases (median: 0.80 [interquartile range (IQR): 0.56–0.81] vs median: 0.05 [IQR: 0.02–0.13], $p\approx 0.0$ from Wilcoxon rank-sum test). The overall AUROC of our model is 0.9448. Considering the small prevalence, PPV and sensitivity are more important for positive detections.
%Our model has a higher-PPV (89.59\%) with lower-sensitivity (67.91\%) indicating that a number of positive samples are missed but the likelihood of the positive prediction being correct is high. 

% PPV and sensitivity are more important and our analysis focus more on the postive cases

\begin{figure}[!t]
\centering
    \begin{subfigure}[t]{0.32\textwidth}%
      \includegraphics[width=\linewidth]{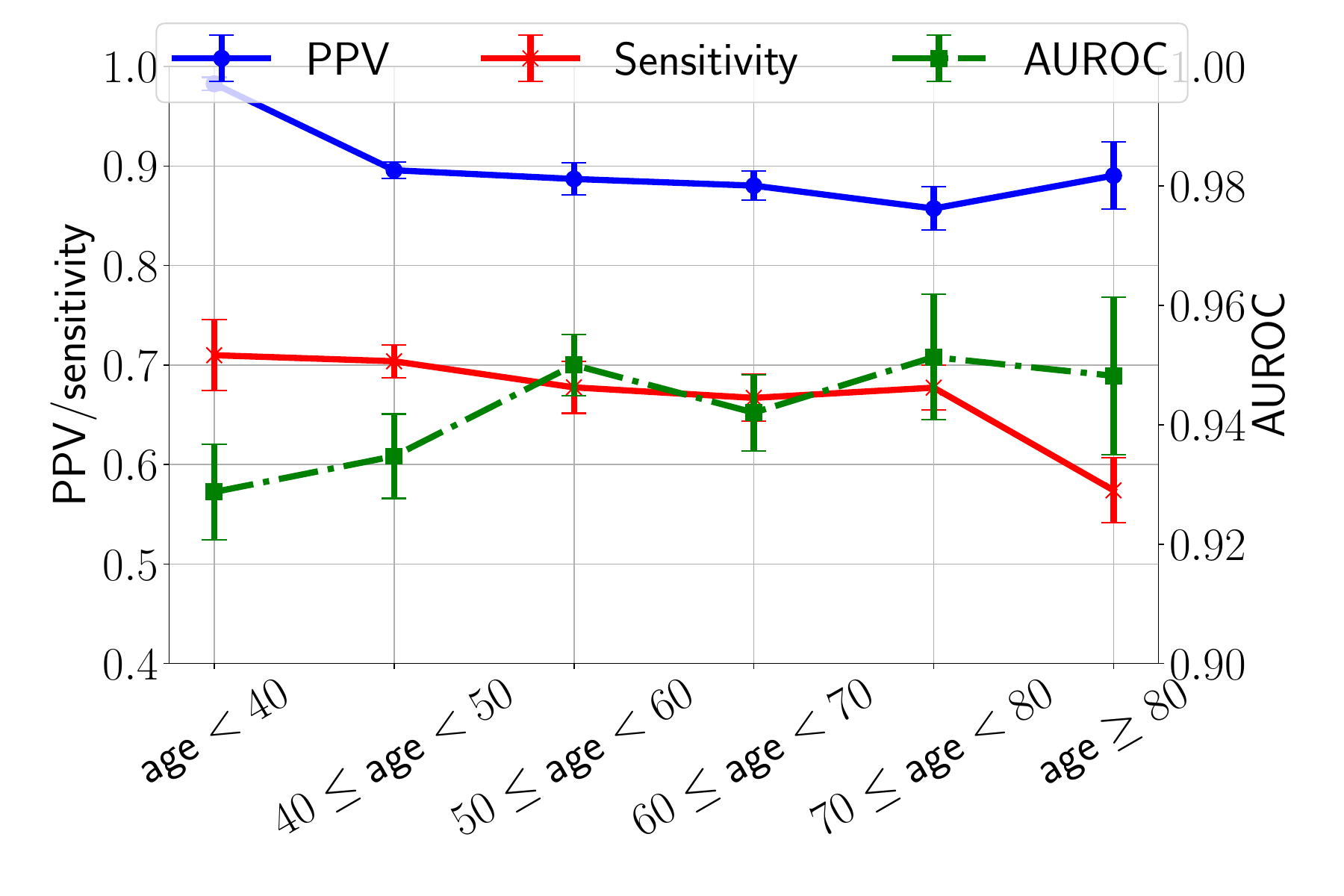}\caption{EMBED-age group}\label{fig:per_age}
      \end{subfigure}
    \begin{subfigure}[t]{0.32\textwidth}%
      \includegraphics[width=\linewidth]{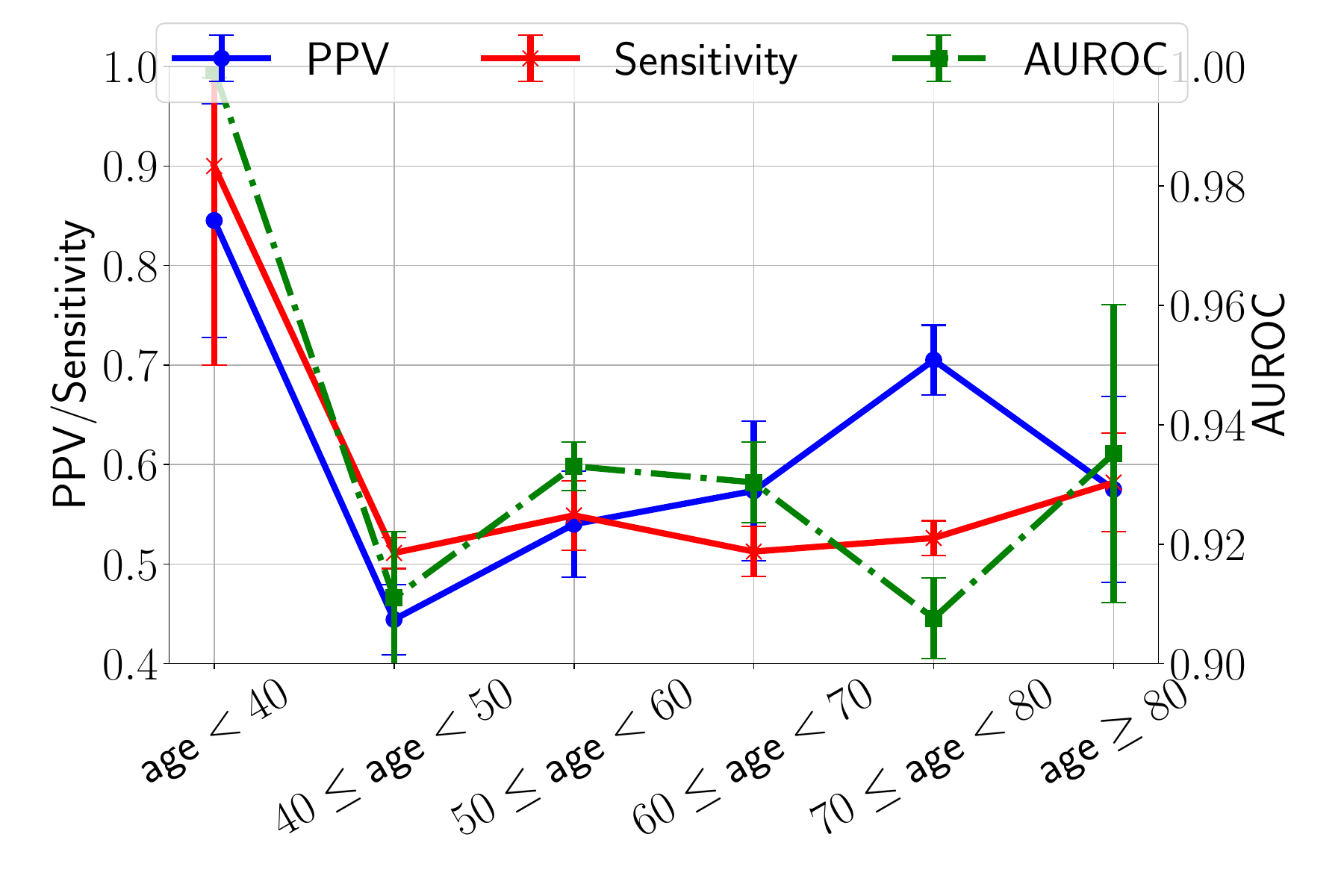}\caption{RSNA-age group}\label{fig:per_age_rsna}
      \end{subfigure}
    \begin{subfigure}[t]{0.32\textwidth}
      \includegraphics[width=\linewidth]{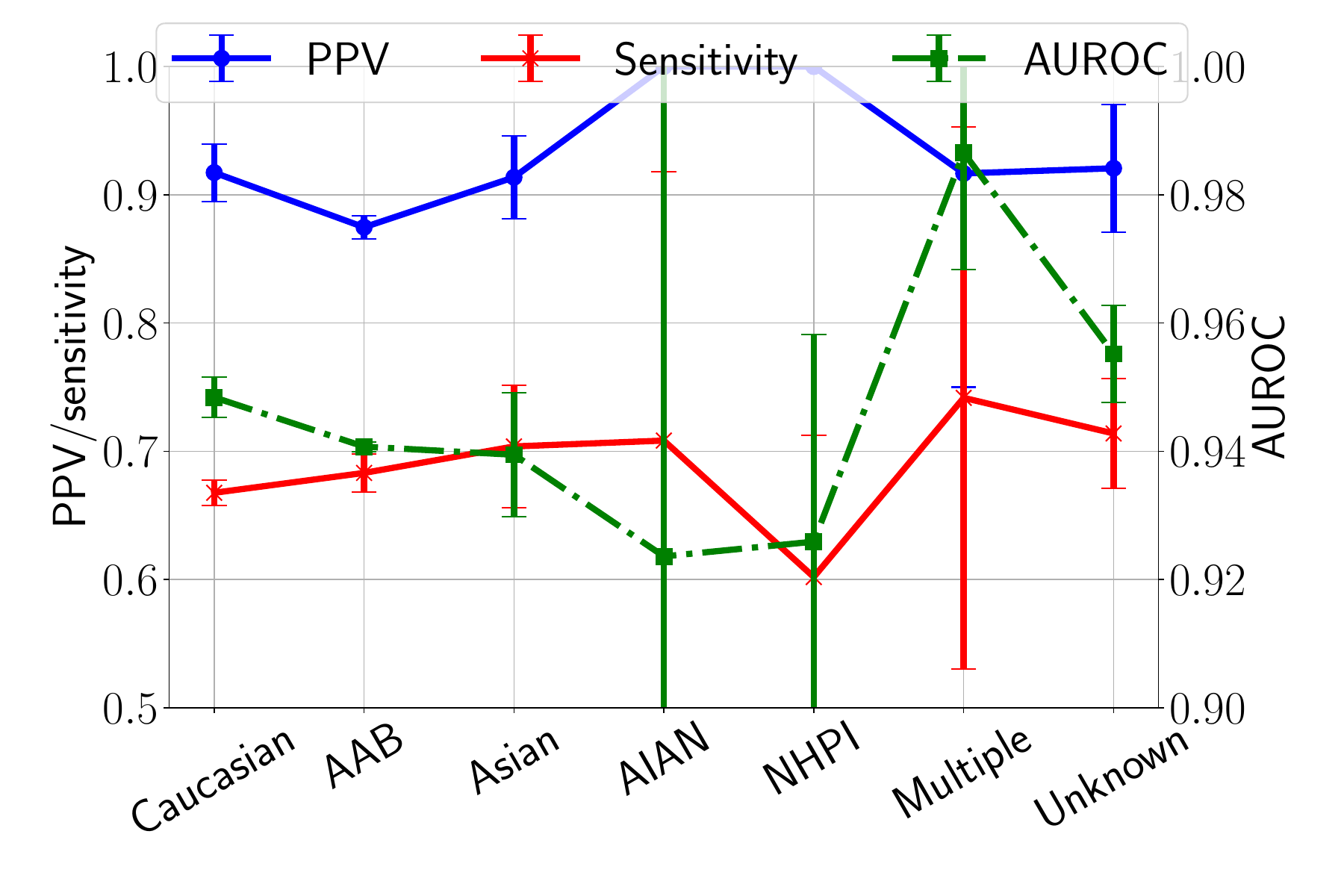}\caption{EMBED-race}\label{fig:per_eth}
      \end{subfigure}\\
    \begin{subfigure}[t]{0.32\textwidth}
      \includegraphics[width=\linewidth]{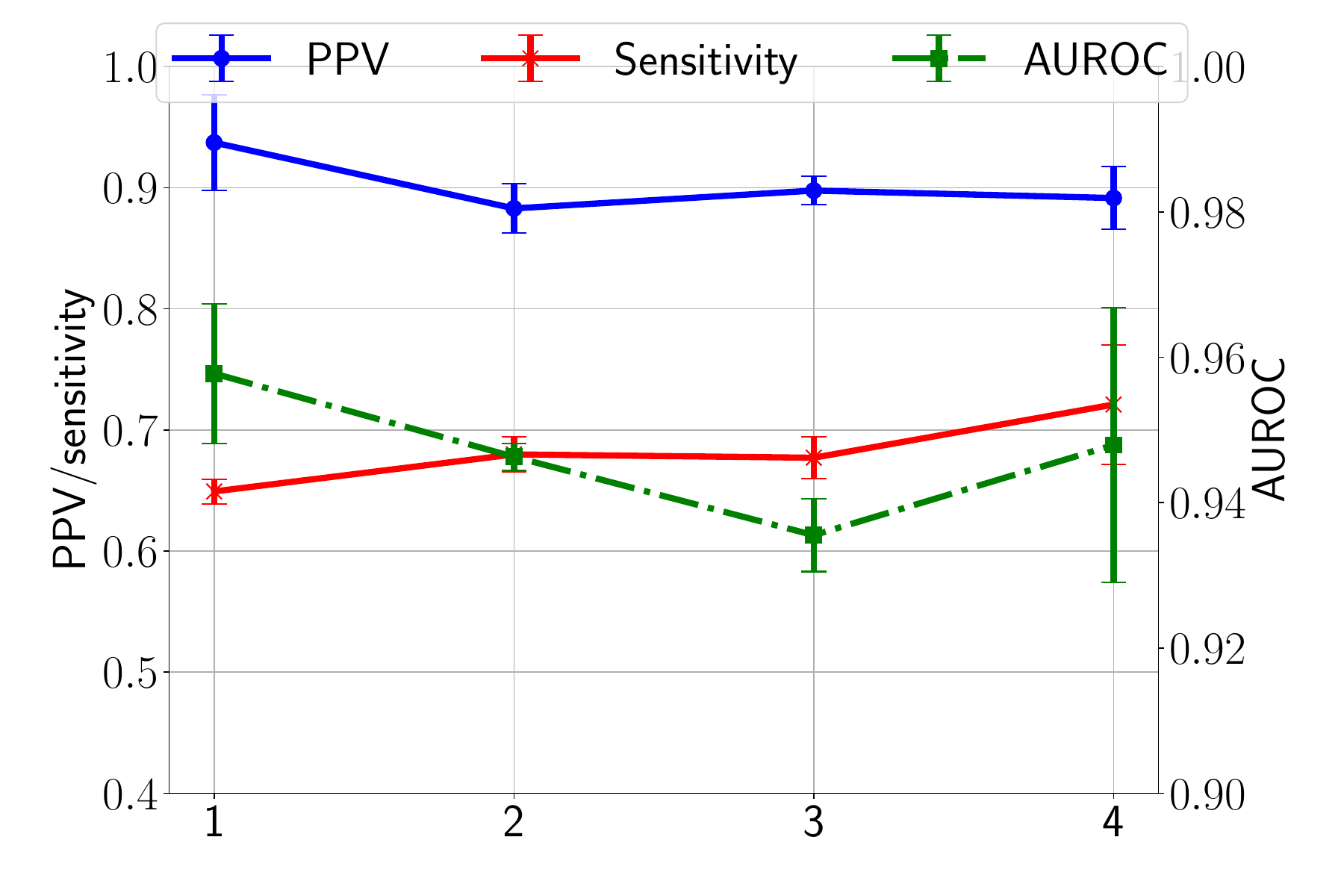}\caption{EMBED-density}\label{fig:per_tisden}
      \end{subfigure}
    \begin{subfigure}[t]{0.32\textwidth}
      \includegraphics[width=\linewidth]{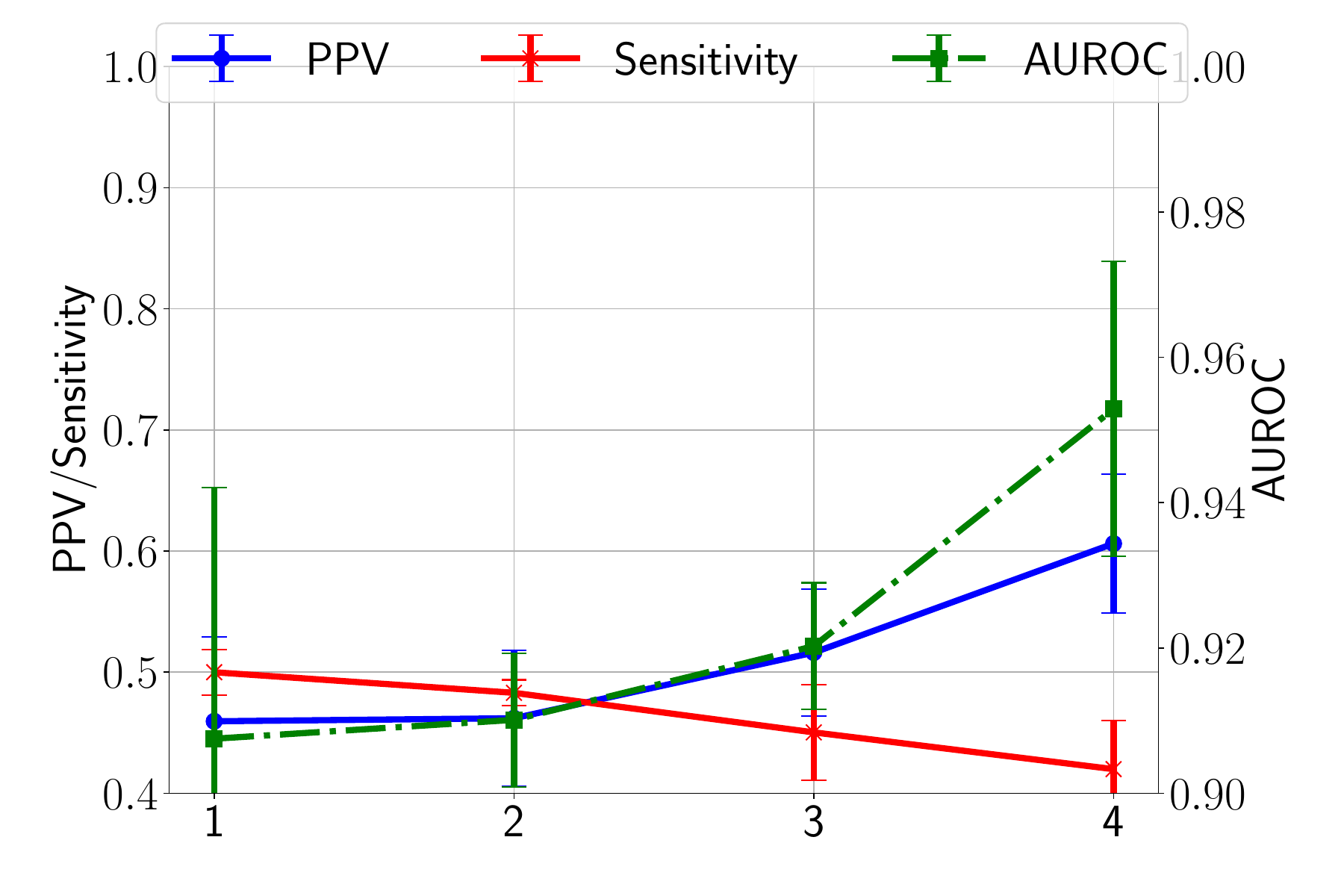}\caption{RSNA-density}\label{fig:per_tisden_RSNA}
      \end{subfigure}
    \begin{subfigure}[t]{0.32\textwidth}
      \includegraphics[width=\linewidth]{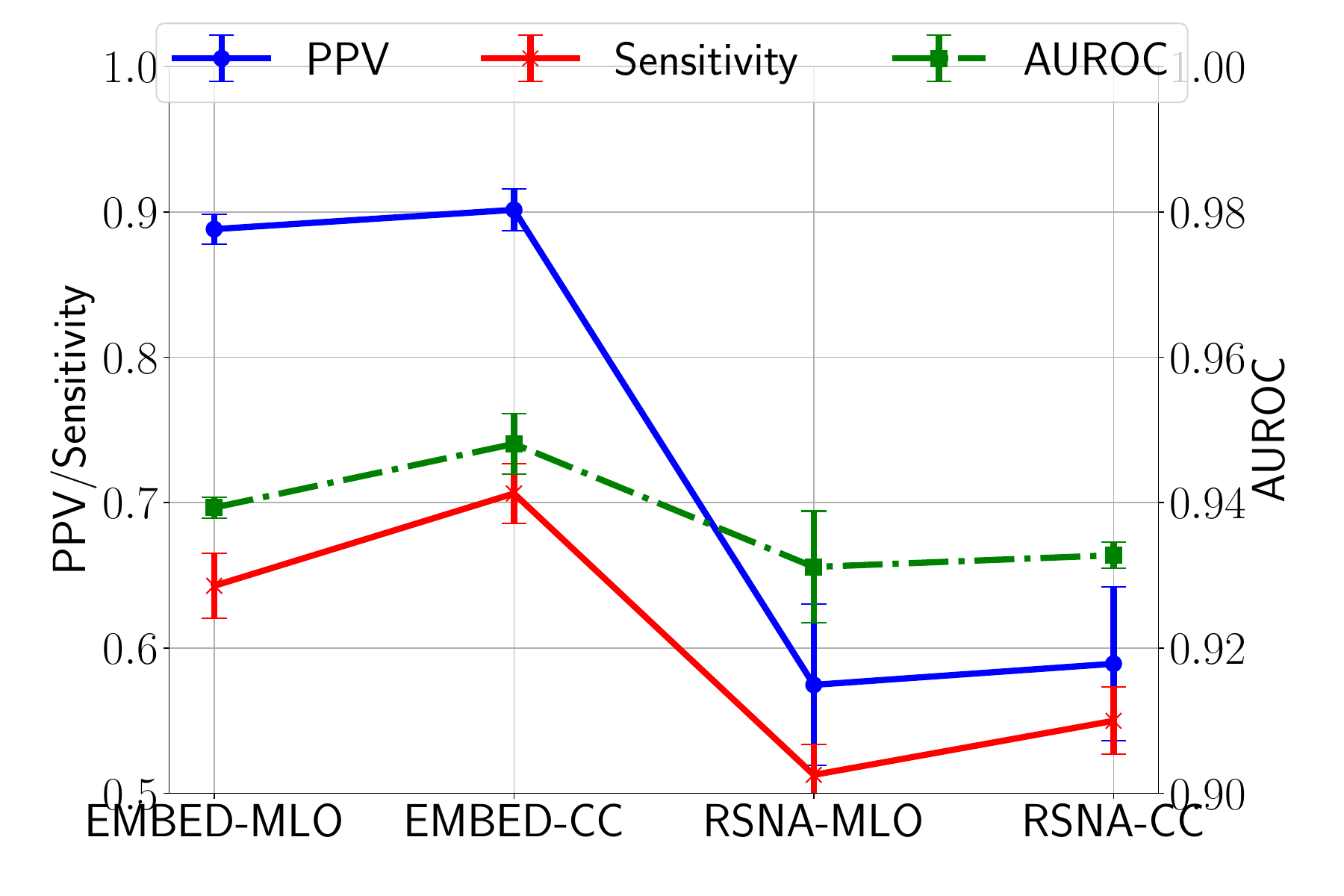}\caption{view position}\label{fig:per_view}
      \end{subfigure}\\
    \begin{subfigure}[t]{0.32\textwidth}
      \includegraphics[width=\linewidth]{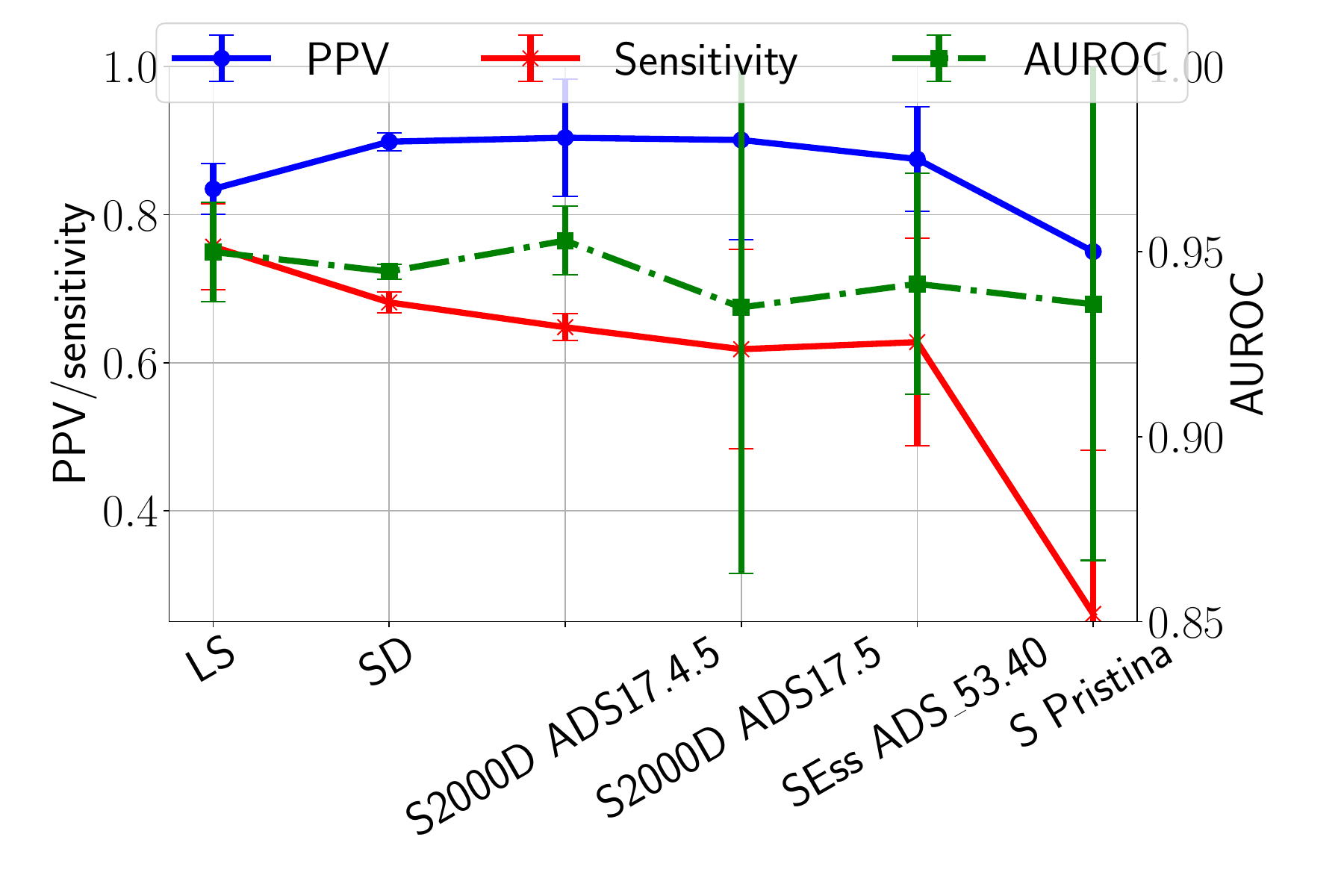}\caption{EMBED-scanner}\label{fig:per_machine}
      \end{subfigure}
    \begin{subfigure}[t]{0.32\textwidth}
      \includegraphics[width=\linewidth]{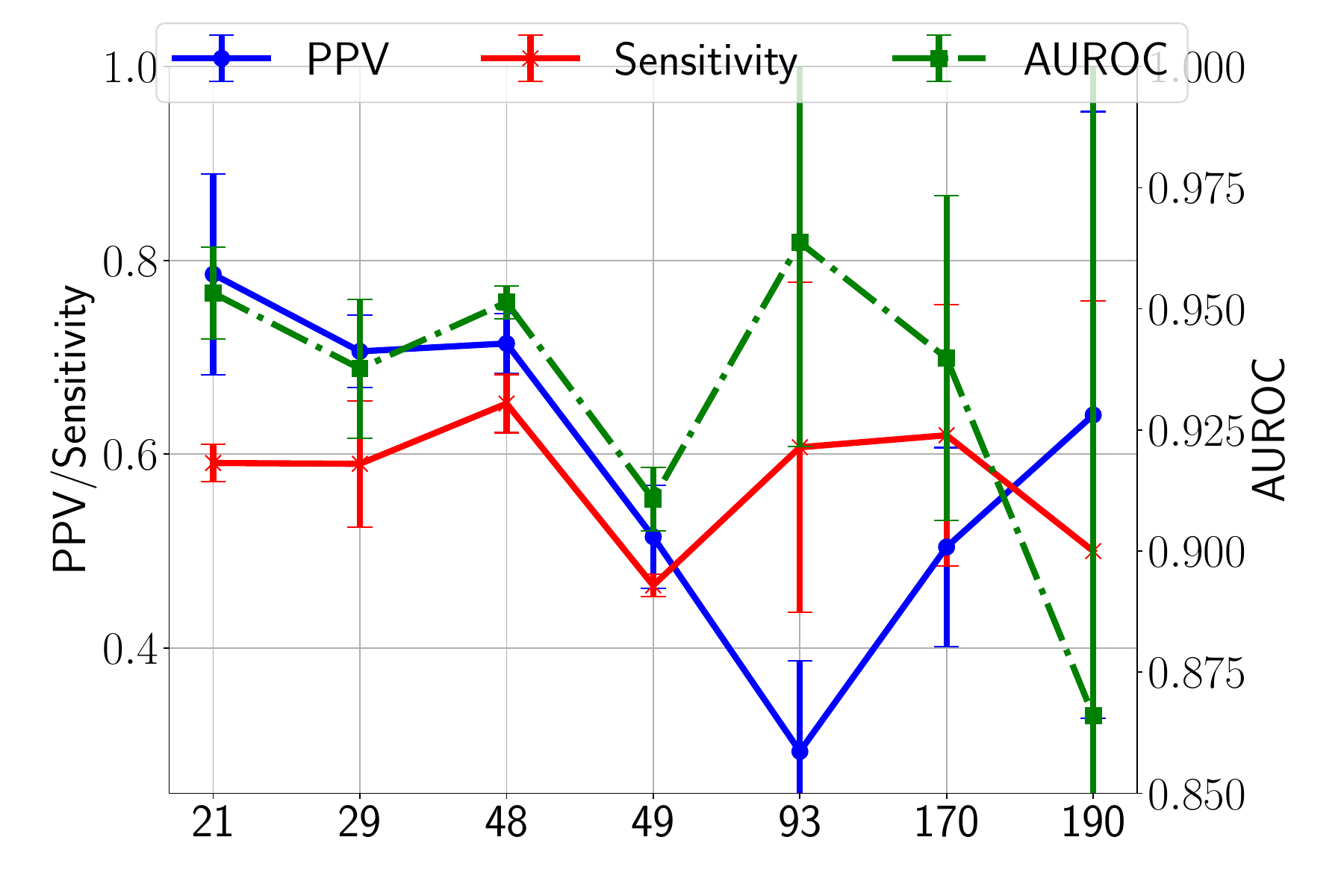}\caption{RSNA-machine ID}\label{fig:per_machine_RSNA}
      \end{subfigure}
    \begin{subfigure}[t]{0.32\textwidth}
      \includegraphics[width=\linewidth]{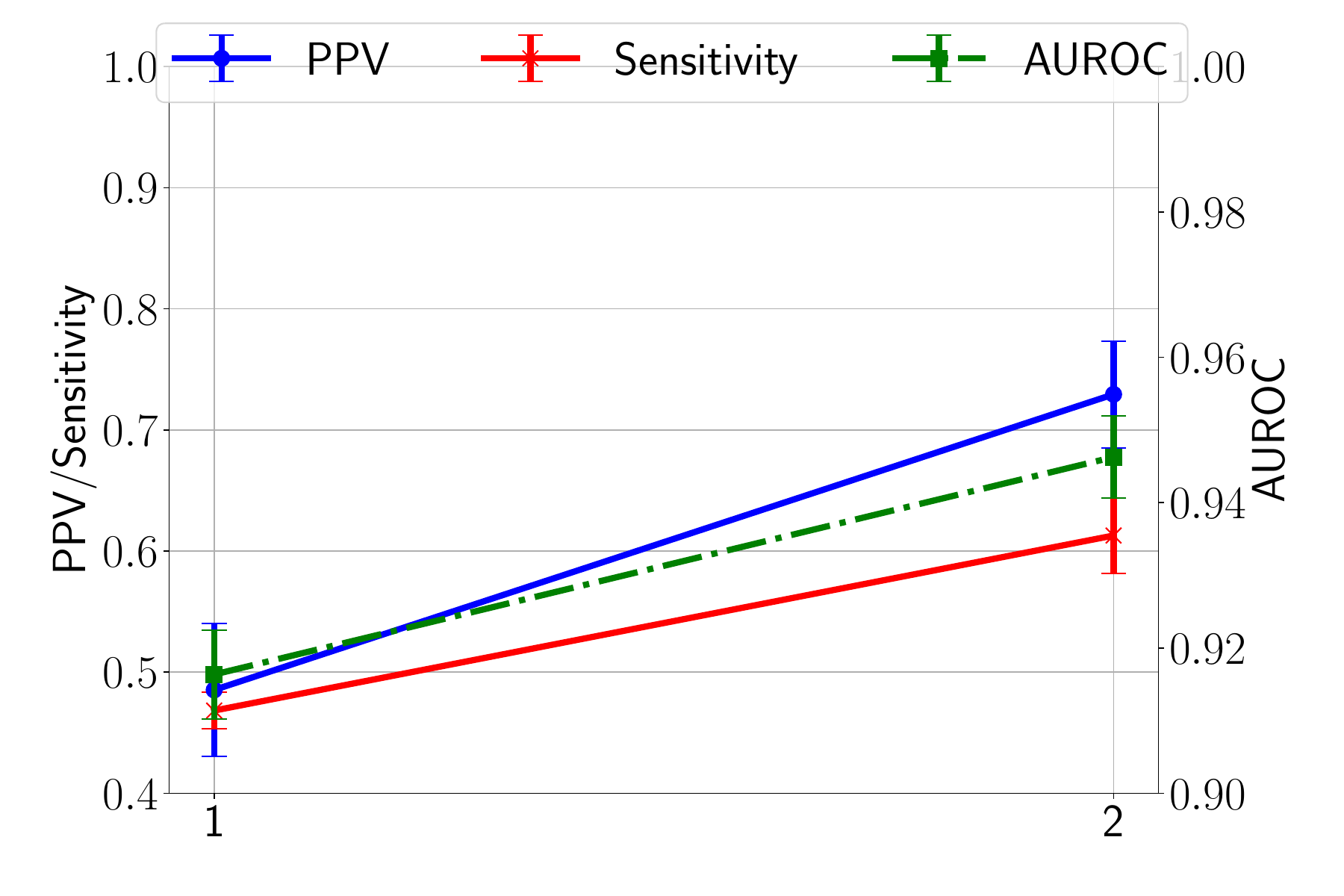}\caption{RSNA-site}\label{fig:per_site}
      \end{subfigure}\\
\caption{
  %Subgroup distribution (top row) with 
  Model performance for different attributes. PPV and sensitivity are shown on the left axis, while AUROC is shown on the right axis. Error bars represent the standard deviation of metrics.}\label{fig:subgroup_pop}
\end{figure}

\subsection{Subgroup Analysis}\label{sub_results}
We utilize the pretrained model and our trained model to perform subgroup analysis on RSNA and EMBED, respectively.
We discuss model performance in relation to the data distribution across subgroups defined by the following attributes: age; race; breast tissue density; view positions; site; and scanning technology. We present our results using the metrics: PPV, sensitivity and AUROC and list all the metrics in Appendix \ref{apd:first}: Table \ref{tab:detailed_EMBED} and Table \ref{tab:detailed_RSNA}. The objective of this analysis is to determine if there are performance disparities affecting specific subgroups, even within models that generally perform well, and to explore the implications of these disparities. For all attributes, we reference Table \ref{tab:num_samples} to examine the distribution of subgroups in our discussion.

\subsubsection{Age} 
Results for different age groups are shown in Figs. \ref{fig:per_age} and \ref{fig:per_age_rsna}. 
For EMBED, participants of age $< 40$ has a high prevalence of 28.14\%. After this, the prevalence gradually decreases from 4\% to 2.34\% with increasing age range except for the group with age $\ge 80$ when it goes up to 4.78\%. One potential explanation for the high prevalence of participants of age $< 40$ is the selection bias - for instance, oversampling of high-risk younger patients, rather than routine screening cohorts. This skew may inflate the model's apparent performance in younger patients and may not reflect a typical population distribution.

We observe that sensitivity declines with increasing age group (coefficient, $\rho=-0.8380, p< 0.04$). In particular, we observe a significant drop in sensitivity for the group with age $\ge80$ than $ \text{age} < 50$ ($p=0.02$). To be specific, the Caucasian population with age $\ge80$ has a lower sensitivity (51.27\%) despite the fact that the overall sensitivity of Caucasian population is not impacted by this drop (see Section \ref{sec:race}). While subgroup-specific threshold recalibration for this age group increased sensitivity by 4.71\% with a minor drop in specificity (by 0.41\%), this improvement remained well-below the overall average sensitivity.
It is to be noted that the U.S. Preventive Services Task Force recommends routine screening from 40 to 74 years of age (\cite{nicholson2024screening}), so our observed performance decline in patients for age $\ge 80$ raises questions about the model’s generalizability beyond recommended screening intervals. Future work could explore specialized or alternative imaging strategies for this age group.
Higher PPV is achieved for the group with age $<40$ than the `$70 \le \text{age} < 80$' subgroup ($p<0.004$). A significant correlation is found between the prevalence and the PPV ($\rho=0.9668,p<0.002$) for the age groups. AUROC is lower for the `age $<40$' subgroup than the `$70 \le \text{age} < 80$' subgroup and it increases with increasing age group (coefficient, $\rho=0.8145, p< 0.05$). 

% 0.6212 (4.71\% increase) sensitivity for age > 80 after subgroup specific recalibration of threshold with a specificity of 0.9923 (0.41\% drop).

For RSNA, prevalence increases with increasing age group from 0.90\% to 4.97\%. Sensitivity is higher for the `age $< 40$' group than the `$40 \le \text{age} < 50$' ($p=0.02$) and `$60 \le \text{age} < 70$' ($p=0.04$) subgroups. However, there are only a few number of positive cases for age $<40$. PPV is lower for the `$40 \le \text{age} < 50$' group than the `age $< 40$' ($p=0.002$) and `$70 \le \text{age} < 80$' ($p=0.04$) subgroups. Moreover, AUROC is lower for the `$40 \le \text{age} < 50$' ($p<0.02$) and `$70 \le \text{age} < 80$' ($p=0.003$) subgroups than the `age $< 40$' group.
%No significant correlation is found between the metrics and the prevalence or age groups for RSNA. 

\subsubsection{Race}\label{sec:race}
Table \ref{tab:num_samples} and Fig. \ref{fig:per_eth} present the distribution and performance based on different racial groups in EMBED. No racial information is found for RSNA.
A higher prevalence is found for the African American or Black (AAB) (3.84\%), Native Hawaiian and Pacific Islanders (NHPI) (4.01\%), and American Indian and Alaskan Native (AIAN) (4.95\%) groups than the overall prevalence.
%The model scored a lower precision for the African American population ($p<0.05$) and higher precision for the NHPI and AIAN groups than the rest of the population ($p<0.05$). Similarly, a drop in recall is observed for the NHPI group though it does not appear to be significant in the ANNOVA test. 
The model scored a lower PPV for the AAB population ($p<0.007$). Here, most mammograms of AAB population are from the Selenia Dimensions scanner (Section \ref{sec:machine}) with a PPV of 87.29\%.
Similarly, a drop in sensitivity is observed for the NHPI group though it does not appear to be significant in the test. 
The high variability in sensitivity with only few positive samples for the AIAN, NHPI, and Multiple groups indicate that the sensitivity estimates might be unreliable for these groups. No significant difference is found among the AUROC scores of racial groups ($p>0.05$). 
%A negative correlation is found between the prevalence and the AUROC ($\rho=-0.8185, p<0.03$) of racial groups.

\subsubsection{Tissue Density}
The subgroup distribution and the evaluation metrics based on breast tissue density are presented in Table \ref{tab:num_samples}, and Figs. \ref{fig:per_tisden} and \ref{fig:per_tisden_RSNA}. 
The definitions of tissue densities are presented in Table \ref{tisden_def} in Appendix \ref{birads_def}. 
We ignore the male tissue density (density of 5). For EMBED, we observe that prevalence increases with tissue density ($\rho=0.9955,p\approx0.05$), especially a high prevalence (5.48\%) is found for density of 4.
Sensitivity is higher for the density of 4 than density of 1 ($p<0.02$). Specifically, AAB population with tissue density of 1 has a lower sensitivity (62.05\%) though it is not reflected in the overall AAB sensitivity score.
No significant difference is found among the PPV and AUROC scores ($p> 0.05$).
%Model uncertainty increases with the tissue density ($\rho=0.9969, p=0.003$).

% as the targets are different.

For RSNA, the model achieves higher AUROC for density of 4 than density of 2 ($p<0.03$). No significant difference is found among the PPV scores. Sensitivity decreases with the tissue density ($\rho=-0.9919, p=0.008$). The observed lower sensitivity for RSNA in denser breast categories (BIRADS C or D) aligns with clinical experience, where extremely dense breast tissue can mask malignancies. Consequently, patients with higher densities may benefit from alternative or supplemental screening (e.g., ultrasound, MRI, or DBT). 
However, similar results may not be expected from the EMBED model as the target variable is different.

\subsubsection{View Position}
The distribution and performance metrics based on MLO and CC view positions are shown in Table \ref{tab:num_samples}, Fig. \ref{fig:per_view}. 
For EMBED, even though the model achieves similar PPV for both view positions ($p=0.3429>0.05$), there is a significant drop in sensitivity by 6.35\% and in AUROC by 0.88\% ($p<0.03$) for the MLO view position. 
MLO projection is clinically important for visualizing the upper-outer breast. Even a modest performance drop in MLO views might risk overlooking certain lesions. 
Follow-up analysis should evaluate if the model captures morphological cues equally across both MLO and CC, or whether additional MLO-specific training data might be warranted.
However, no significant difference is found in performance between the views for the RSNA dataset. 
%The drop in sensitivity for MLO view in EMBED might happen due to data imbalance.

% scanning tech.
\subsubsection{Scanning Technology}\label{sec:machine}
The distribution and the performance metrics based on different scanners are presented in Table \ref{tab:num_samples}, and Figs. \ref{fig:per_machine} and \ref{fig:per_machine_RSNA}. We observe that for EMBED, most of the samples are from Selenia Dimensions. No significant difference is found in the evaluation metrics among the different machines except that Senograph (S) Pristina achieves a significantly lower average sensitivity ($p<0.007$). This might happen due the low number of samples from this scanner. However, we observed a negative correlation between prevalence and sensitivity scores ($\rho=-0.867, p<0.03$). 

In RSNA, instead of scanner names, machine IDs are provided, where machines 21, 29 and 48 are present at site 2 and the rest are present at site 1. Almost 45\% samples are from machine 49. Similar to EMBED, no significant difference is found in evaluation metrics among the different machines in the RSNA dataset except that machine 93 achieves a significantly lower average PPV ($p<0.02$). However, we observe a negative correlation between prevalence and AUROC scores ($\rho=-0.867, p<0.01$). 

\subsubsection{Sites}
We found information of sites only within the RSNA dataset (presented in Table \ref{tab:num_samples}). The performance metrics based on the two sites are presented in Fig. \ref{fig:per_site}. We observe that for RSNA, site 2 outperforms site 1 in PPV by 24.4\%, in sensitivity by 14.45\% and in AUROC by 3\% ($p<0.03$) despite having more training samples (with higher prevalence) from site 1 than site 2.

\begin{table}[!t]
\centering
\caption{Association ($p<0.05$) of variables with target and predicted label}\label{tab:asso_attribute_target}
\begin{tabular}{l|rrrrr}\hline
&\multicolumn{5}{c}{EMBED} \\\hline
group &age group &race &tissue density &view position &machine ID \\\hline
target &\textbf{0.1754} &0.0242 &\textbf{0.0527} &0.0387 &0.0262 \\
Predicted label &\textbf{0.1441} &0.0256 &\textbf{0.0487} &0.0398 &0.027 \\\hline
&\multicolumn{5}{c}{RSNA} \\\hline
group &age group &site &tissue density &view position &machine ID \\\hline
target &\textbf{0.0836} &0.0099 &0.0175 &$<0.0001$\footnotemark[2] &0.043 \\
Predicted label &\textbf{0.0637} &0.0255 &0.0274 &0.0012\footnotemark[2] &0.035 \\\hline
\end{tabular}
\footnotetext[2]{indicates that the association is not significant ($p>0.05$)}
\end{table}

\vspace{5mm}
In addition to the individual attribute analysis, we also conducted a joint attribute analysis, including age-race, race-density, and age-density for EMBED, as presented in Appendix \ref{apd:first}. The findings from this analysis are consistent with those discussed in this section.\\

From the preceding analysis, the model for EMBED achieves good PPV ($>88\%$) for almost all subgroups except for the `$70 \le \text{age} < 80$' group and the AAB group. The model also achieves a good sensitivity for almost all subgroups ranging from 66.70\% to 74.17\%, except the drops for the Caucasian population with age $\ge 80$, the AAB population with tissue density of 1, NHPI race, and MLO view groups. 
The model for RSNA achieves reasonable performance for all subgroups, except a lower PPV for the `$40 \le \text{age} < 50$' group, and lower sensitivity for the density of 4 group. High AUROC is achieved for most subgroups for both datasets except the drops for `age $<40$' group in the EMBED-AI model and for the `$40 \le \text{age} < 50$' and `$70 \le \text{age} < 80$' groups in RSNA-AI model. The lower performing groups may benefit from alternative or supplemental screening. 
For most cases in both datasets, no significant difference is found among the performance metrics of different scanners. 
% The drop in performance for MLO view in EMBED may be due to the data imbalance.
High specificity ($\geq 98.44\%$ from Table \ref{tab:detailed_EMBED} - \ref{tab:detailed_RSNA}) indicates low false positive rates for all subgroups. Model uncertainty demonstrates only a nominal variation, remaining within a narrow range across the subgroups.
Moreover, we observe that the EMBED-AI model demonstrates lower variability in performance across subgroups compared to the RSNA model. %Therefore, EMBED is a more suitable dataset to train BCD-AI models from digital mammography considering the subgroup performance and data variability in EMBED.

%From the observations discussed above we can conclude that based on the sensitive features, the model does not perform consistently across all the subgroups, which we validate using the fairness measures discussed next. 

To investigate how much additional performance deviation is introduced by the training algorithm or the model, we present the statistical association between the targets (or predicted labels) and the attributes in Table \ref{tab:asso_attribute_target}. The associations between targets and attributes represent the underlying relationships of data, and the associations between attributes and model predictions represent how much of the relationship is mirrored by the model. From the Table we observe that there is a significant association present only between the age groups and targets for both datasets. The model mirrors the relationships significantly well for most attributes except that there is a slight drop in association between age groups and predicted labels. 

\begin{figure}[!t]
\centering
    \begin{subfigure}[t]{0.7\textwidth}
      \includegraphics[width=\linewidth]{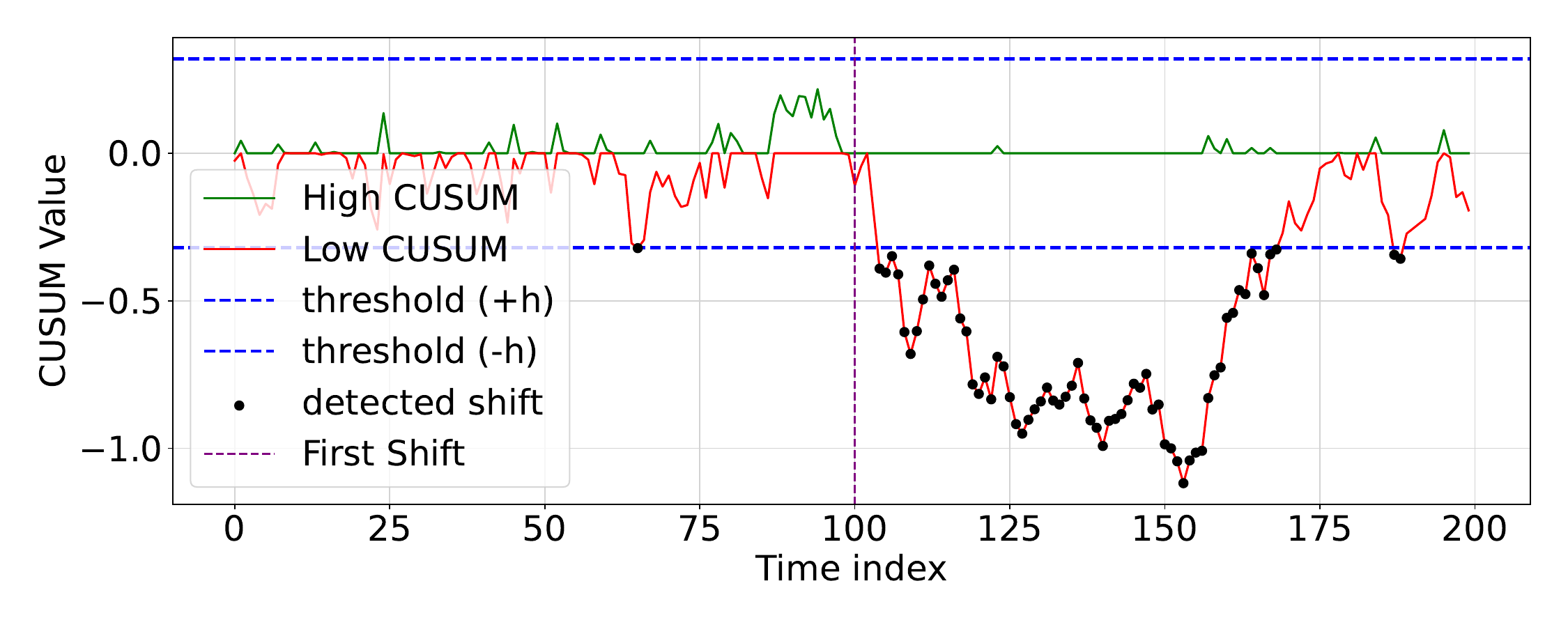}\caption{}\label{fig:moni_eth}
      \end{subfigure}
    \begin{subfigure}[t]{0.7\textwidth}
      \includegraphics[width=\linewidth]{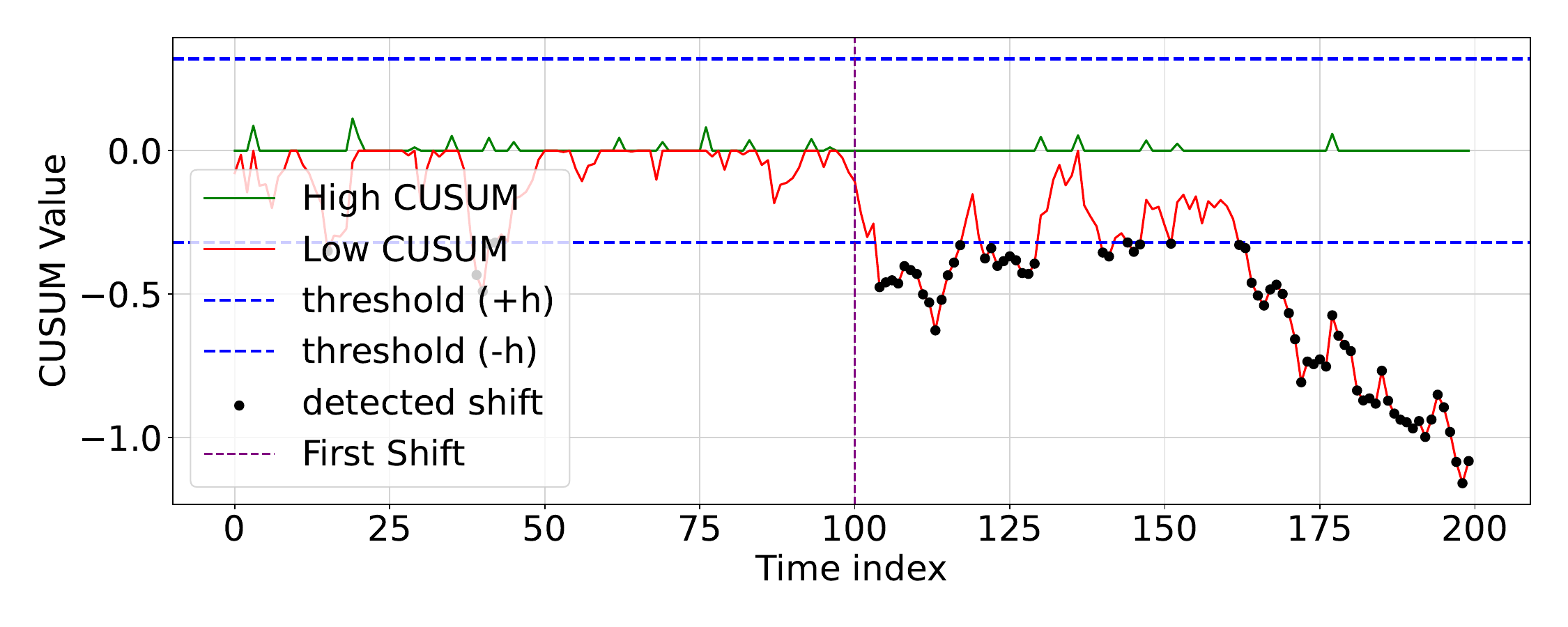}\caption{}\label{fig:moni_age}
      \end{subfigure}
  %  \subfigure[for some untuned $k$]{\label{fig:moni_FP}
  %    \includegraphics[width=0.9\linewidth]{subgroup_figures/CUSUM_FP_fold1_metric_recall_atttissue_density_0.5.pdf}}
\caption{CUSUM based sensitivity monitoring charts. The batches are assumed to form a sequential data. After time index 100, we add more samples from the underperforming (a) the NHPI and (b) `age $\ge 80$' groups of EMBED to each batch to introduce performance drift.}\label{fig:perf_moni1}
\end{figure}

\subsection{Monitoring Performance Drift}
Here, we show how CUSUM can be used to monitor drift in model performance. Specifically, we monitor the sensitivity. The number of bootstrapped batches $N$ is set to 200 and the batch size $B$ is $1000$. The number of positive samples for a subgroup within a batch is set equal to the base prevalence ratio from the data distribution.
We assume that the first 100 batches are formed from the base empirical distribution and the rest of the batches come from the shifted distribution.

To present a specific case, after batch 100, we start adding more samples to each batch from the underperforming NHPI and `age $\ge 80$' subgroups of EMBED at a fixed $\Delta p$ of 0.3 to introduce a drift in sensitivity, which we aim to detect. $k$ is tuned from $\{0.0,0.01,0.02,...,0.1\}$. We present the drift detection charts in Fig. \ref{fig:perf_moni1}. 
FAR and drift detection delays are presented in Table \ref{tab:moni_results} in Appendix \ref{sec:app_perf_moni}. There are only a few false alarms with a minimal average detection delay. 
The few initial false positives in Fig. \ref{fig:perf_moni1} indicate a temporary drop in sensitivity. Then, the control signal goes up again within the threshold range. However, the detected sharp drift after index 100 indicates a permanent degradation in sensitivity below the acceptable range and follow up adjustments to the model is necessary. Therefore, the CUSUM based monitoring method is effective in quickly detecting permanent performance drifts.

%The proposed framework could be adopted on multiple large datasets to perform similar subgroup analysis and an ensemble method could be developed for a general purpose BCD-AI model for deployment and robust performance monitoring in clinical settings.
%Monitoring performance drift using CUSUM is crucial during the deployment of AI models in clinical practice because it enables early detection of potential biases or degradation in model performance as new patient data is introduced. This is particularly important when the demographic distribution of the data shifts, as models may underperform for certain subgroups, leading to unfair or unreliable predictions. By identifying and addressing such drifts in real-time, healthcare providers can ensure that the model continues to perform consistently across diverse patient populations, safeguarding the quality and fairness of medical decisions.

\begin{figure*}[!t]
\centering
    \begin{subfigure}[t]{0.32\textwidth}%
      \includegraphics[width=\linewidth]{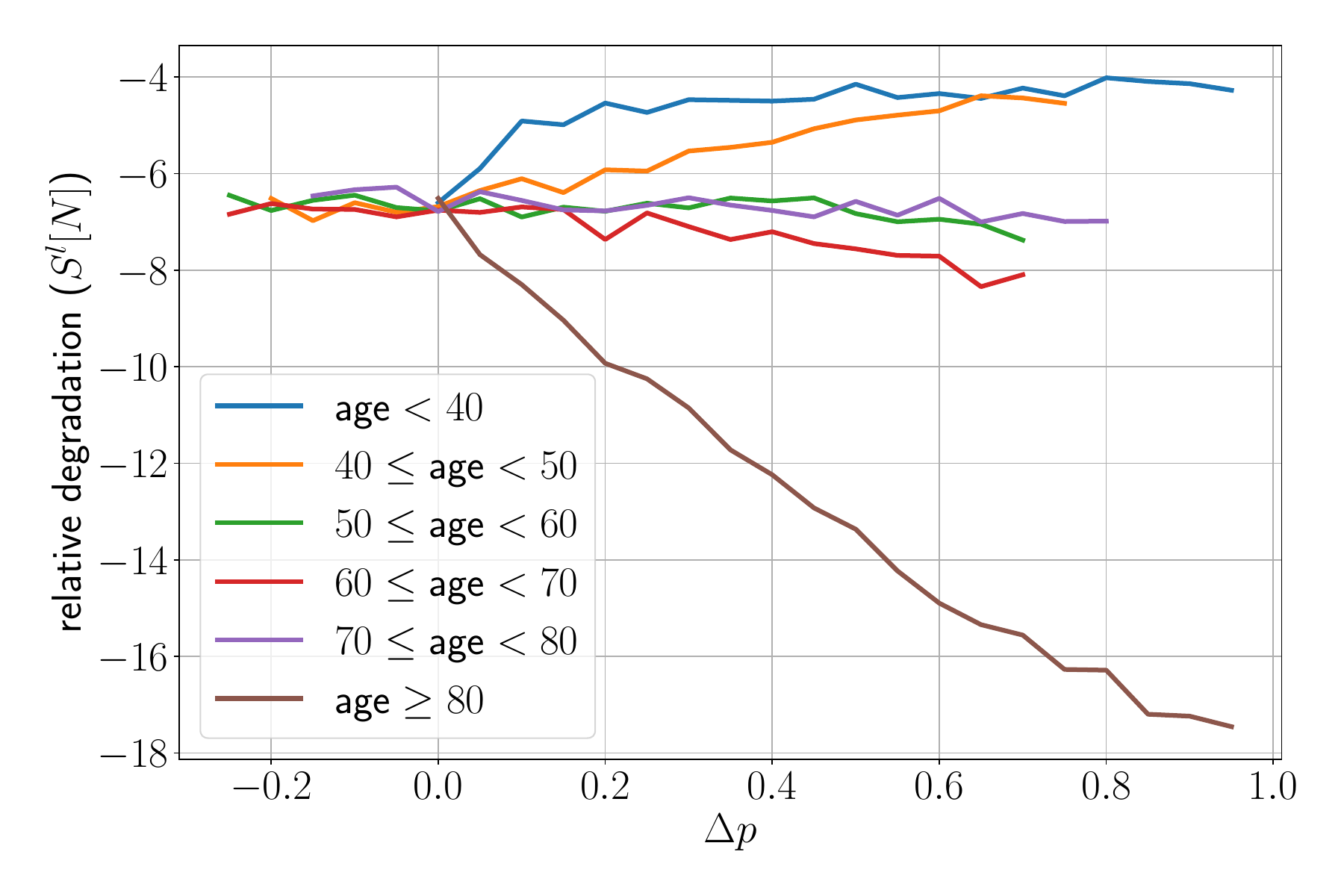}\caption{}\label{fig:monitoring_age_recall}
      \end{subfigure}
    \begin{subfigure}[t]{0.32\textwidth}
      \includegraphics[width=\linewidth]{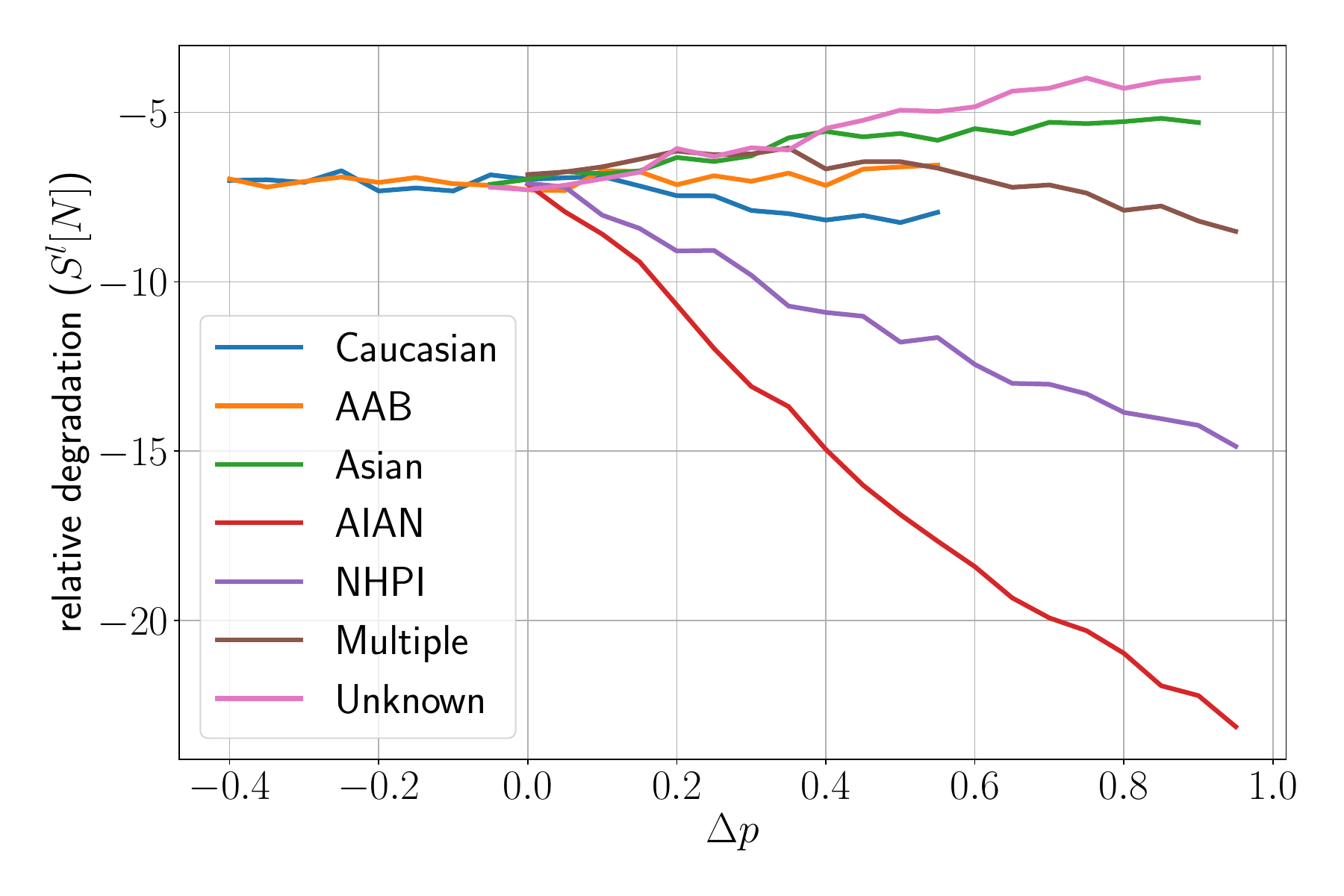}\caption{}\label{fig:monitoring_race_recall}
      \end{subfigure}%\\
    \begin{subfigure}[t]{0.32\textwidth}
      \includegraphics[width=\linewidth]{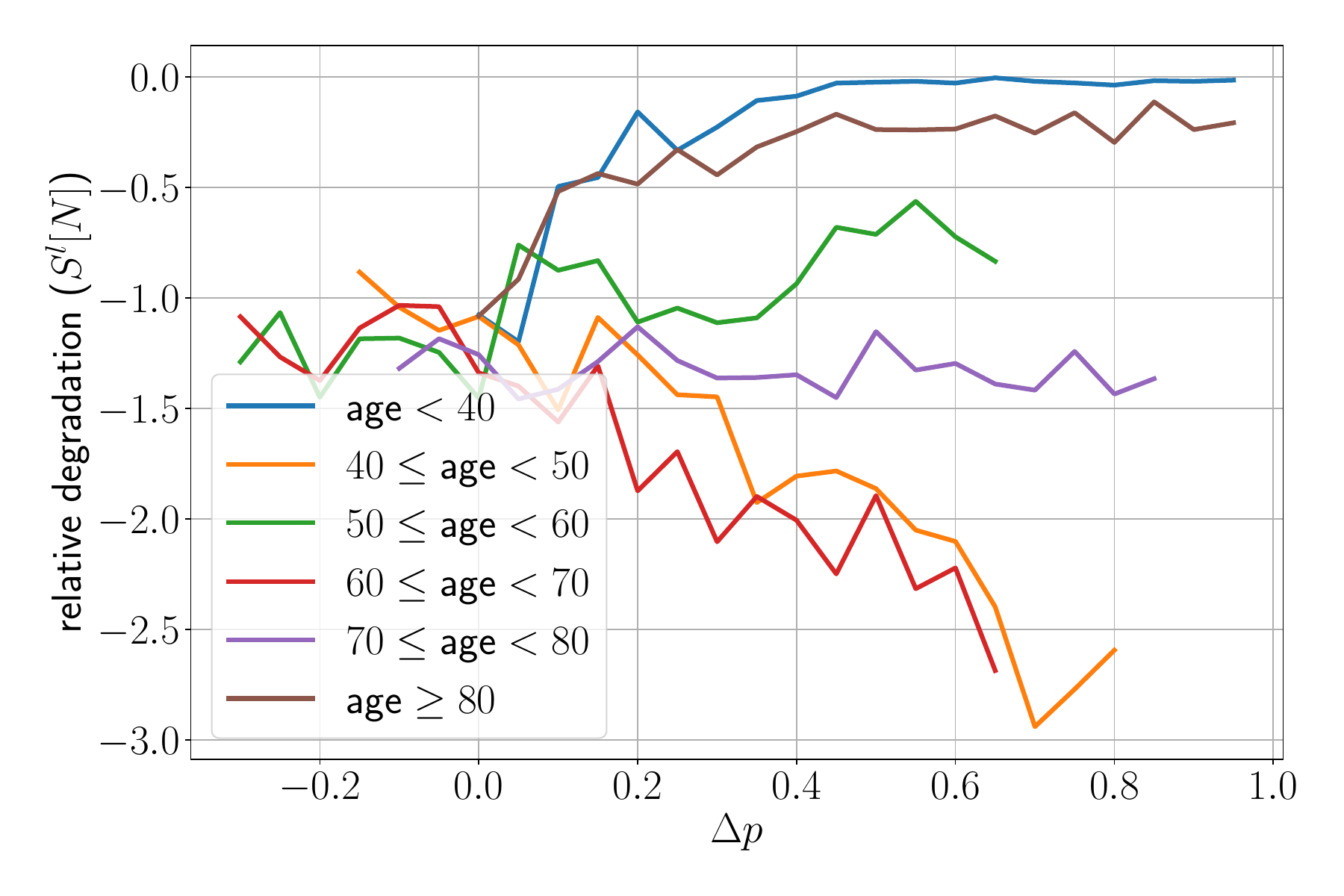}\caption{}\label{fig:monitoring_age_recall_rsna}
      \end{subfigure}%
\caption{
  Monitoring sensitivity drops under the distribution shifts of subgroups, such as by deviating the proportion of (a) age groups from EMBED, (b) racial groups from EMBED and (a) age groups from RSNA. Here, $k$ is set to 0.}\label{fig:monitoring_cusum}
\end{figure*}

Next, we simulate CUSUM monitoring by varying the deviating subgroup $u$ and the deviation $\Delta p$ to monitor the effects of these shifts in data distribution. The lower control value at the end of a certain simulation $S^l(N)$ represents the relative quantity of performance drop for the introduced deviation at the simulation. We present these relative sensitivity drops in Fig. \ref{fig:monitoring_cusum}. 
We observe from Fig. \ref{fig:monitoring_age_recall} that for EMBED, compared to the performance of base distribution (when $\Delta p = 0$), sensitivity drops rapidly even when there is a slight increase of samples from the `age $\ge 80$' group. However, for the `$60 \le \text{age} < 70$' group, the sensitivity drops slowly with large deviations. Similarly, in Fig. \ref{fig:monitoring_race_recall}, we observe that sensitivity drops at different rates with increasing deviations for the AIAN and the NHPI group. On the contrary in Fig. \ref{fig:monitoring_age_recall_rsna}, although we observe sensitivity drops for large deviations in certain age groups, the drops are small for the RSNA dataset compared to the baseline performance.

Monitoring performance drift with CUSUM is important, as it enables the early detection of performance degradation when new data are introduced, particularly when demographic shifts lead to underperformance. By identifying these changes in real time, CUSUM ensures that model performance remains accurate, equitable, and reliable, helping to prevent bias and ensuring fair treatment across diverse patient populations. This proactive monitoring not only helps sustain the effectiveness of predictive models during deployment but also mitigates the risk of worsening outcomes for underrepresented subgroups.

%Monitoring performance drift with CUSUM is vital in clinical practice, as it allows for early detection of performance degradation when new data are introduced, especially if demographic shifts cause models to underperform for certain subgroups. Real-time detection ensures consistent, fair, and reliable model performance across diverse patient populations.

\section{Related Work}\label{related_work}
Since the introduction of deep learning, computer aided diagnosis received a great deal of attention in the research community. BCD models for digital mammography can ease the screening process for practitioners and reduce mortality. Researchers explored various deep architectures and methodologies to train BCD models. \cite{sahu2023recent,raiaan2024mammo,qureshi2024breast,mahoro2022applying,petrini2022breast,wang2023evaluating,altameem2022breast,shen2019deep} train BCD models on different datasets. However, these models are of limited capacity as they are trained on smaller datasets with degraded resolutions and only a few positive cases. For example, the datasets such as Mini-DDSM by \cite{lekamlage2020mini,heath1998current}, CMMD by \cite{cui2021chinese}, CDD-CESM by \cite{khaled2021categorized}, BMCD by \cite{loizidou2021digital} have less than 10000 mammograms, while VinDr-Mammo by \cite{pham2022vindr} has 20000 mammograms. OMI-DB by \cite{halling2020optimam} is a large dataset in mammography. However, these datasets lack sufficient patient demographic and other clinical information, which raises concerns about the possibility of bias in the previous models.  

In this regard, EMBED by \cite{jeong2023emory} and RSNA (\cite{rsna-breast-cancer-detection}) are large datasets that include rich information about demographics, and clinical attributes. We find only a few works in the literature that utilized these datasets for developing BCD AI models. 
\cite{khara2024generalisable} train ResNets for tissue density and race classification, while
\cite{donnelly2024asymmirai} develop a 1 to 5 year cancer risk prediction model on EMBED. 
\cite{correa2024efficient} designed an adversarial debiasing approach with partial learning to reduce racial disparities for the density classification task.
\cite{hwang2023impact} train ResNet models for BCD with only 1441 samples from EMBED and the other datasets and assessed AUROC across the racial groups.
\cite{zhang2023multivariate} develop abnormal lesion classification models and analyze the performance across subgroups based on demography, pathology outcomes and image findings. These works select a balanced dataset for training, which does not reflect the real scenario. Therefore, there is a need for a BCD model trained on a large and diverse dataset and for analyzing the model across the most important subgroups.

\cite{sun2022performance} demonstrate a comprehensive performance analysis on the COVID-19 detection model from chest X-rays. \cite{afzal2023towards} perform a comprehensive analysis for kidney tumor segmentation across the demographic groups. \cite{mehta2024evaluating} analyze the uncertainty in model performance among the subgroups for three medical applications. Following this, we aim to analyze our BCD AI model performance across the relevant subgroups. Recently, \cite{nguyen2024patient} studies the false positive rates across different demographic groups on a commercial AI model for the DBT examinations. However, it is hard to understand the significance of these results since no details are provided about the training data.

\cite{vela2022temporal} demonstrates that the quality of AI model may drop over time for various applications. For BCD-AI models, the performance of the model may drop due to shifts in the distribution of demographics, scanning technology, and local processes (\cite{sahiner2023data}). In this regard, \cite{nestor2019feature} demonstrates that the performance of the model trained on electronic health records degrades due to distribution changes. To monitor the performance over time, \cite{prathapan2024quantifying, zamzmi2024out} adopt methods from statistical process control either using simulated mean shifts in performance or by inserting out of distribution data. However, in our paper, we monitor the model performance due to shifts in patient demographics.

\section{Conclusion}\label{conclusion}
We develop a high-performing AI model on EMBED using advanced computer vision training strategies using limited computational resources. 
%We train it on the diverse EMBED dataset, leveraging its demographic and clinical attributes. We then assess performance across subgroups to identify biases. 
%While the models are trained on diverse datasets, leveraging the demographic and clinical attributes and performs well overall, we observe a drop in performance metrics in a few cases.
Despite training the models on diverse datasets that incorporate demographic and clinical information and achieving overall good performance, we identify a few significant cases of performance disparities. For EMBED, we observe lower PPV for the `$70 \le \text{age} < 80$' and African American groups, and lower sensitivity for the Caucasian population with `age $\ge 80$', the African American population with tissue density 1, NHPI population and MLO view groups. 
For RSNA, we observe reasonably good performance for all subgroups, except a lower PPV for the `$40 \le \text{age} < 50$' group, and lower sensitivity for the density of 4 group.
Unreliable sensitivity estimates of AIAN and NHPI population due to low number of samples emphasizes the need to collect more data for these populations to ensure fairness for all patients. 
Existing data balancing or targeted fairness-driven retraining techniques may not significantly improve the sensitivity estimates due low number of positive test samples for these groups.
Our analysis reveals that our EMBED model exhibits lower performance variability across subgroups compared to the RSNA model. 
% Our Analysis reveals that EMBED is better suited for training BCD-AI models than RSNA.
%Fairness metrics indicate greater unfairness towards the demographic subgroups. 
With this subgroup analysis framework, we can identify the biases inherited in the training data and show that biases in the data could lead to potential disparities in clinical outcomes for some groups even with well-trained AI models. Care should be taken while deploying such models by monitoring the performance of such underperforming subgroups.
Finally, we show that CUSUM based charts can effectively monitor the model performance over time as local populations shift (e.g. new age demographics, variations in scanners, or locations), where it can quickly flag the performance drop, enabling appropriate interventions, such as model retraining or recalibration. The results from the comprehensive analysis can inform the development of evaluation techniques for unbiased breast cancer detection models from digital mammography.

Implementing a monitoring system that flags performance drops for underrepresented groups can have several clinical implications. One of the key benefits is the timely identification of performance disparities. CUSUM allows for real-time monitoring and quickly detecting when a model's accuracy diminishes for specific demographics, such as certain age groups, racial or ethnic communities. Early identification of these disparities enables clinicians to take action, ensuring that underserved groups are not left at the disadvantage of a lower prediction of BCD. Further, as local populations evolve, whether through demographic shifts or changes in healthcare infrastructure (e.g., variations in scanner types or staff training), the performance of BCD models can also fluctuate. A continuous monitoring method helps address these changes by quickly flagging any decline in model performance due to these factors. When such a performance drop is detected, it can prompt timely interventions like model retraining with updated, representative data or recalibration to restore performance. This dynamic approach ensures that models remain effective as healthcare conditions evolve. This real-time monitoring can also enable proactive clinical interventions. For example, if a performance issue is flagged for an underrepresented group, healthcare providers can adjust the model’s parameters or incorporate additional training data to improve predictions. Such interventions enhance the model’s ability to detect breast cancer more accurately across all patient groups, especially in communities that may have historically faced challenges with underdiagnosis or misdiagnosis. This proactive approach ensures that the model continues to serve all populations effectively, potentially reducing health disparities in breast cancer detection. 

In the long-term, the use of continuous monitoring can contribute to a more equitable healthcare system. By continuously refining models based on real-world performance data, biases can be identified and corrected early, preventing the reinforcement of healthcare disparities. This creates a cycle of continuous improvement, where BCD models evolve to become more inclusive and accurate for all populations, ensuring that advancements in technology benefit everyone, regardless of demographic factors.

\backmatter

\bmhead{Author Contribution}

\bmhead{Funding}

\bmhead{Data Availability}
The datasets used in this research are found publicly available at \cite{rsna-breast-cancer-detection} and \cite{jeong2023emory}. 
%The project codes will be available soon. 
%\href{https://anonymous.4open.science/r/BCD_on_EMBED-0761/README.md}{here}\footnote{\href{https://anonymous.4open.science/r/BCD_on_EMBED-0761/README.md}{anonymous.4open.science/r/BCD\_on\_EMBED-0761/}}.

\bmhead{Acknowledgments}
%The authors would like to thank X.
We are grateful of the support and resources provided by the  University of Maryland Institute for Health Computing and for the IT support provided by Jonathan Heagerty throughout the project.
%Please refer to Journal-level guidance for any specific requirements.

\section*{Declarations}
\bmhead{Ethical Approval}
This research does not require Institutional Review Board approval.
\bmhead{Conflict of Interest}
The authors declare no competing interests.

\bibliography{EMBED_comprehensive_study}% common bib file

\newpage
\begin{appendices}
%\section{BIRADS Definitions, and Challenge Dataset}
\section{BIRADS Definitions}\label{birads_def}
\begin{figure}[!h]
\centering
\includegraphics[width=0.8\linewidth]{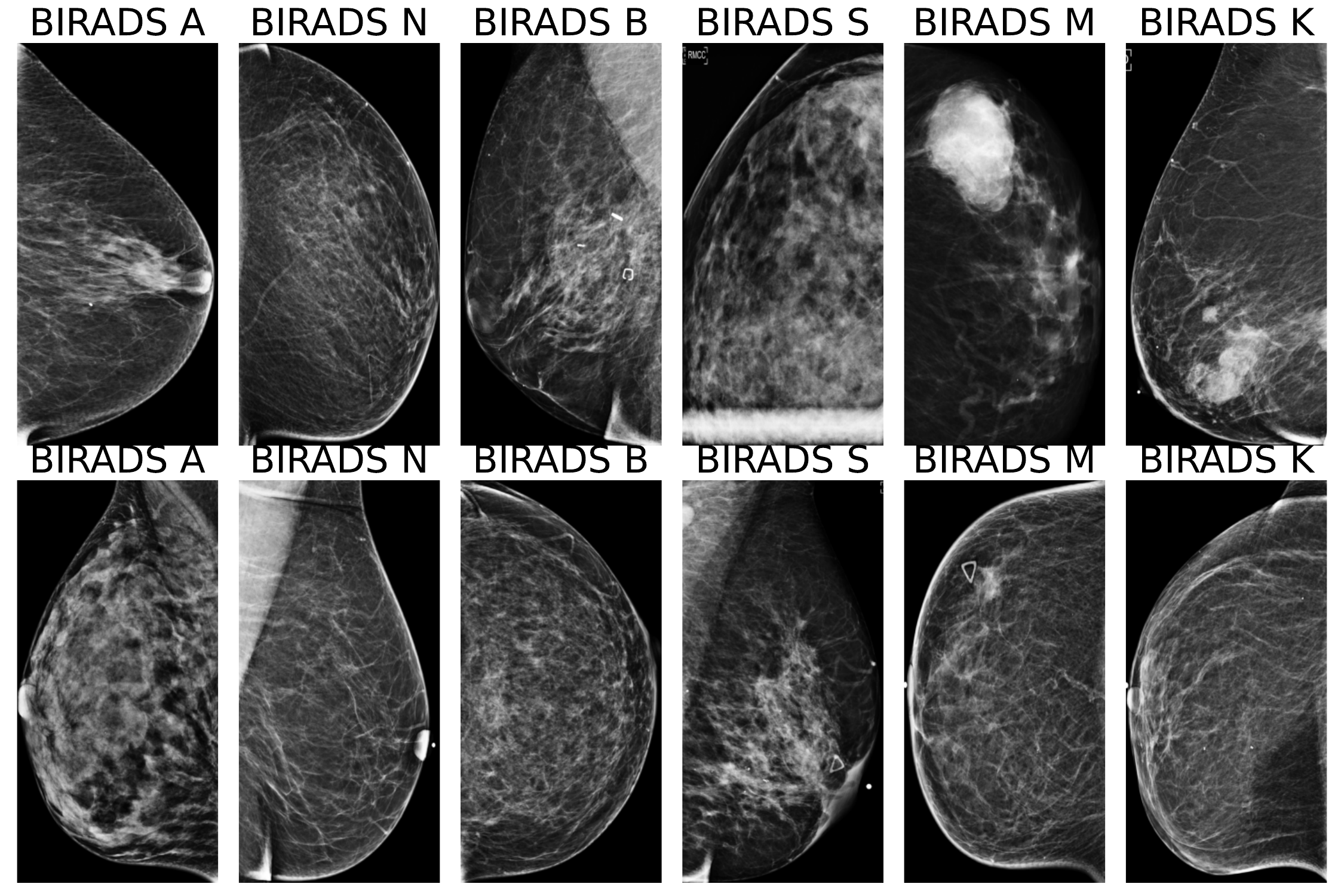}
\caption{Sample mammograms for different BIRADS scores}\label{fig:BIRADS_samples}
\end{figure}

We provide the definitions of different BIRADS scores in Table \ref{BIRADS_definition} and tissue densities in Table \ref{tisden_def}. 
%The number of samples of each BIRADS score in a fold of the 4 fold cross-validation are presented in Table \ref{num_samples}, while 
Some mammogram examples based on the BIRADS scores are presented in Fig. \ref{fig:BIRADS_samples}. 

\begin{table}[h]
\caption{Definition of BIRADS tissue densities}
   \begin{tabular*}{0.6\textwidth}{cccc}
    \hline 
    Density & BIRADS && Tissue description \\ \hline 
1 & A && fatty tissue \\
2 & B && scattered fibroglandular \\
3 & C && heterogeneously dense \\
4 & D && extremely dense \\
5 & -- && male tissue \\\hline
\end{tabular*}%
  \label{tisden_def}%
\end{table}%

\begin{table*}[h]\centering
\caption{Definition of BIRADS scores for mammogram screening}
   \begin{tabular}{p{1.3cm}|p{1.6cm}|p{1.1cm}|p{1.1cm}|p{1.4cm}|p{1.1cm}|p{1.2cm}|p{1.2cm}}
    \hline 
    scores & 0 or A & 1 or N & 2 or B & 3 or P & 4 or S & 5 or M & 6 or K \\ \hline 
Findings & additional evaluation required & normal tissue & benign findings & probably benign & sus- picious & highly suspicious & proven malignancy \\ \hline 
\end{tabular}%
  \label{BIRADS_definition}%
\end{table*}%

%\begin{figure*}[!t]
%\floatconts
%  {fig:subgroup_pop_comp}
%  {\caption{Data distribution of different attributes in the Kaggle challenge dataset (\cite{rsna-breast-cancer-detection}). The number of samples are shown on the left axis at a log scale. Prevalence is shown on the right axis at a linear scale.}}
%  {%
%    \subfigure[age group]{\label{fig:comp_pop_age}%
%      \includegraphics[width=0.34\linewidth]{subgroup_figures/Comp_Population_age.pdf}}
%          \subfigure[tissue density]{\label{fig:comp_pop_tisden}
%      \includegraphics[width=0.30\linewidth]{subgroup_figures/Comp_Population_tisden.pdf}} 
%    \subfigure[view position]{\label{fig:comp_pop_view}
%      \includegraphics[width=0.30\linewidth]{subgroup_figures/Comp_Population_view.pdf}}
%  }
%\end{figure*}

\section{Training Details}\label{app:train_details}
Here we present the preprocessing details of each mammogram, the set of input augmentations and the choice of hyperparameters for training the model.

\begin{table}[!htp]\centering
\caption{Correlation or Association ($p<0.05$) between variables within the datasets}\label{tab:data_corr}
\begin{tabular}{l|rrrr}\hline
&\multicolumn{4}{c}{EMBED} \\\hline
group&age&race &density &view\\\hline
race &0.0904 &&& \\
density &-0.21\footnotemark[4] &0.1199 & &\\
view&0.0092 &0.0069\footnotemark[3] &0.008 & \\
machine&0.02449 &0.0538 &0.0509 &0.0063\footnotemark[3] \\\hline
& \multicolumn{4}{c}{RSNA} \\\hline
group&age &site &density &view\\\hline
site &0.2976 &&& \\
density &-0.258\footnotemark[4] & && \\
view&0.0125\footnotemark[3] &0.0087 &0.0292 & \\
machine&0.1426 &1 &0.0652 &0.0163\footnotemark[3] \\\hline
\end{tabular}
\footnotetext[3]{indicates where correlation or association is not significant ($p>0.05$)} 
\footnotetext[4]{instead of considering as categorical variables, here the spearman correlation coefficient is measured between age and density.} 
\end{table}

%\subsection{Description of the Challenge Dataset}\label{Challenge_dataset}
%We present the distribution of attributes, such age participant age, breast tissue density and mammogram view from the challenge dataset by \cite{rsna-breast-cancer-detection} in Fig. \ref{fig:subgroup_pop_comp}. We did not find any details for the racial or ethnicity features. There are 1158 cancer and 53548 normal mammograms in the dataset. It consists of only screening mammograms. Considering the number of samples and the prevalence for the subgroups, we can conclude the challenge dataset distribution differs significantly than the EMBED dataset. 

\subsection{Preprocessing}
For each mammogram, we first read the DICOM meta data, convert it to 8 bit integers, and normalize them between 0 to 255. 
Next, an off-the-shelf YOLOX-nano $416 \times 416$ model (\cite{ge2021yolox}) is applied as the ROI extractor that crops the breast area to retrain the texture of the mammogram. Finally, the image is resized to $1024 \times 2048$ to be utilized as the input to the model. This preprocessing step removes the unnecessary dark background from the input.

\subsection{Augmentations}\label{sec:aug}
We apply a specific set of augmentations to the inputs following \cite{dangnh0611}. During training, we apply random cropping, changes in brightness or contrast, down scaling, geometric distortions such as shifts, rotation, elastic transformations, grid distortions, and random erasures with certain probabilities. We also randomly apply horizontal and vertical flipping with probability of 0.5. We normalize the images with the ImageNet dataset mean and standard deviation. 

\subsection{Hyperparameters}
We train the model with a batch size of 8 for 15 epochs. We use a SGD optimizer with a learning rate of 0.001, momentum of 0.9, weight decay of $2 \times 10^{-6}$ and a cosine annealed scheduling. An exponential moving average (EMA) model is tracked with a decay of 0.9998, and used for final evaluation. A binary cross-entropy loss is employed with smoothing only for the positive samples using a smoothing parameter of 0.2. We use a balanced batch sampling with positive to negative sample ratio at $1:1$. We train the small ConvNext model with dropout at a rate of 0.5 and and a drop path rate of 0.2.

\subsection{Model Uncertainty Estimation}
Model uncertainty is measured using the Monte Carlo (MC) dropout method by \cite{gal2016dropout}. Since our model is trained with dropout, this method involves activating dropout during inference. The model is run multiple times with different dropout masks, generating a distribution of predictions for the same sample. The uncertainty is then defined as the standard deviation of these predictions.

%\newpage
\section{Additional Results}
\begin{figure}[!b]
\centering
    \begin{subfigure}[t]{0.32\textwidth}%
      \includegraphics[width=\linewidth]{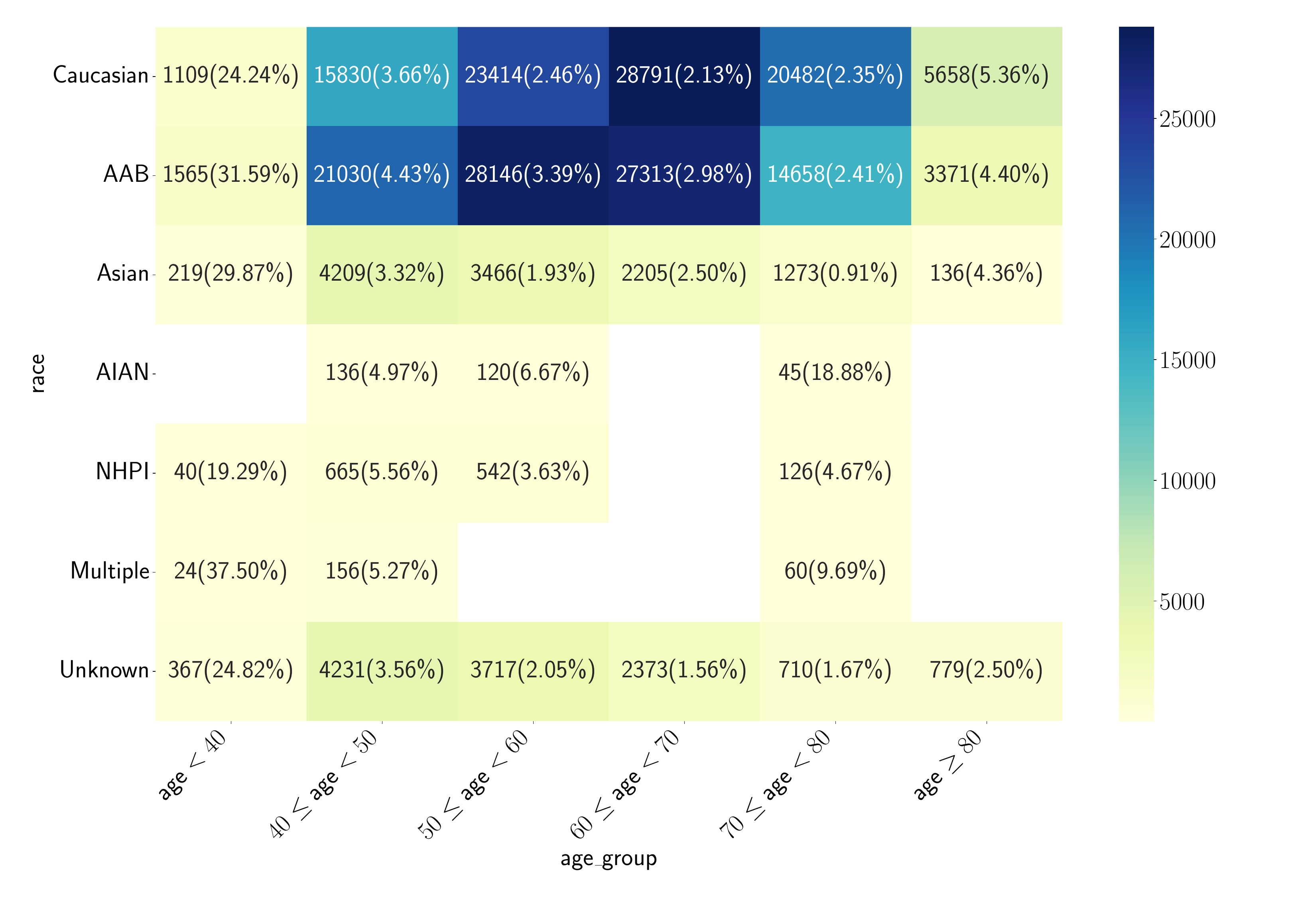}
      \end{subfigure}
    \begin{subfigure}[t]{0.32\textwidth}
      \includegraphics[width=\linewidth]{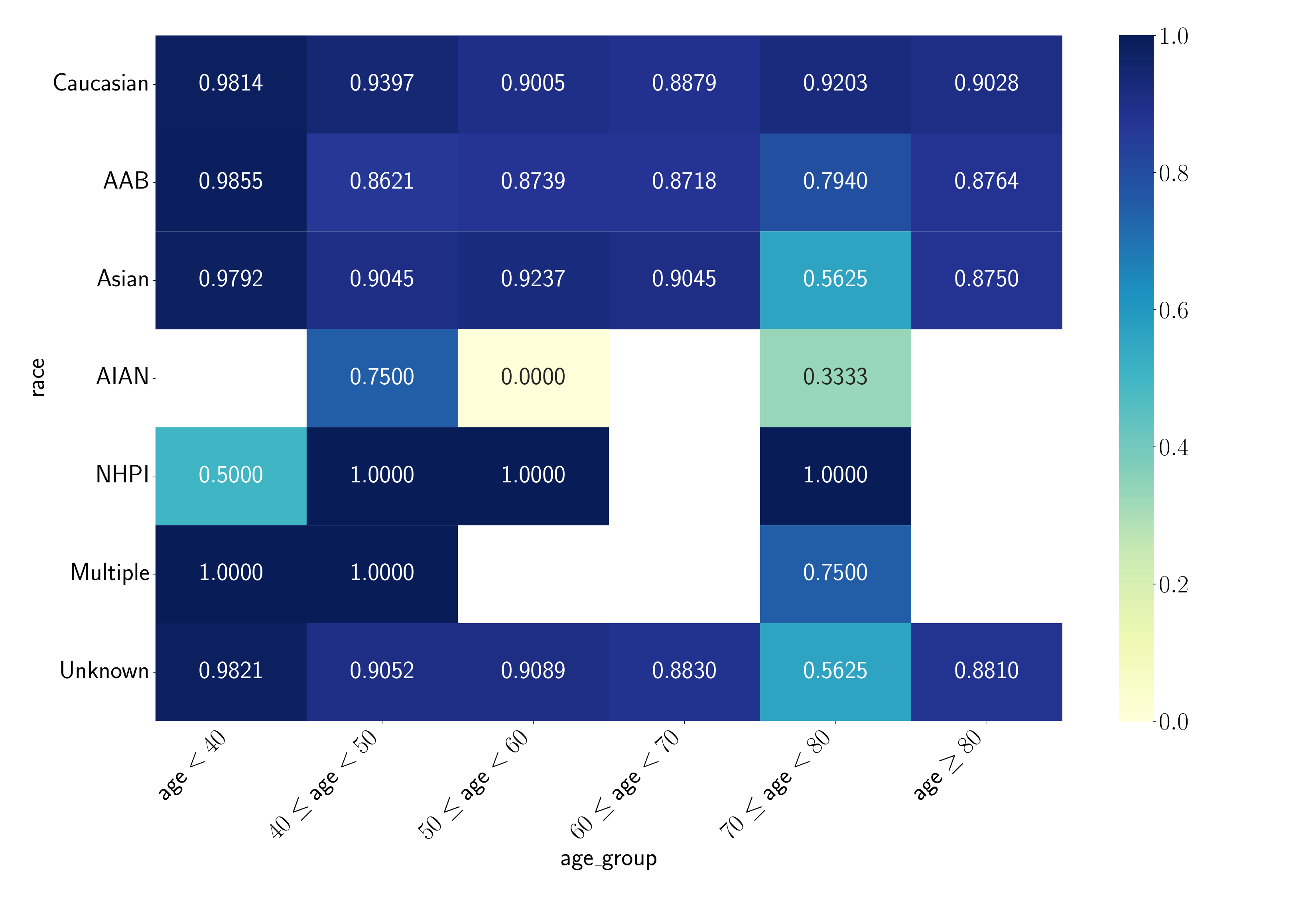}\caption{age group-race}
      \end{subfigure}
    \begin{subfigure}[t]{0.32\textwidth}
      \includegraphics[width=\linewidth]{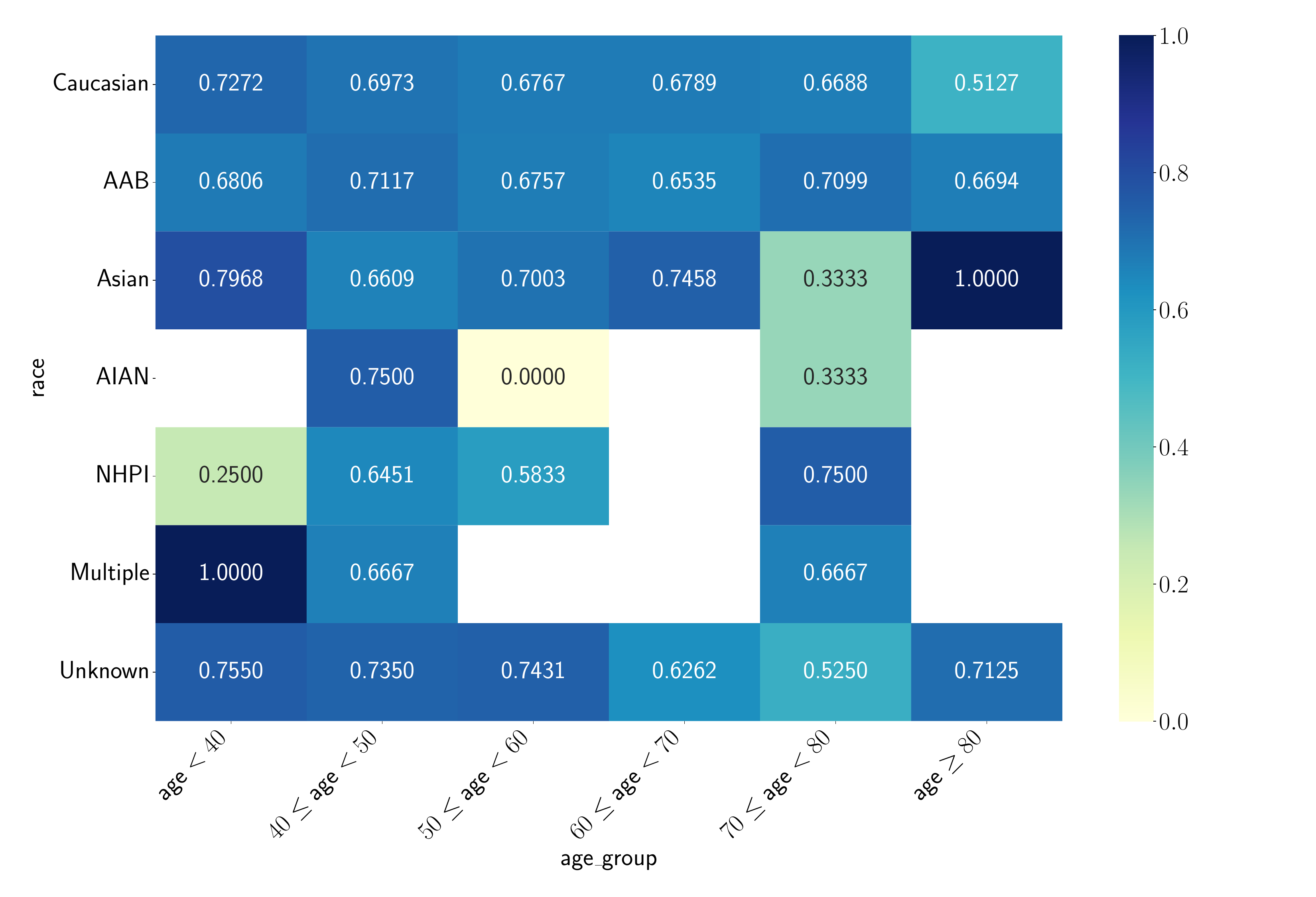}
      \end{subfigure}\\
    \begin{subfigure}[t]{0.32\textwidth}%
      \includegraphics[width=\linewidth]{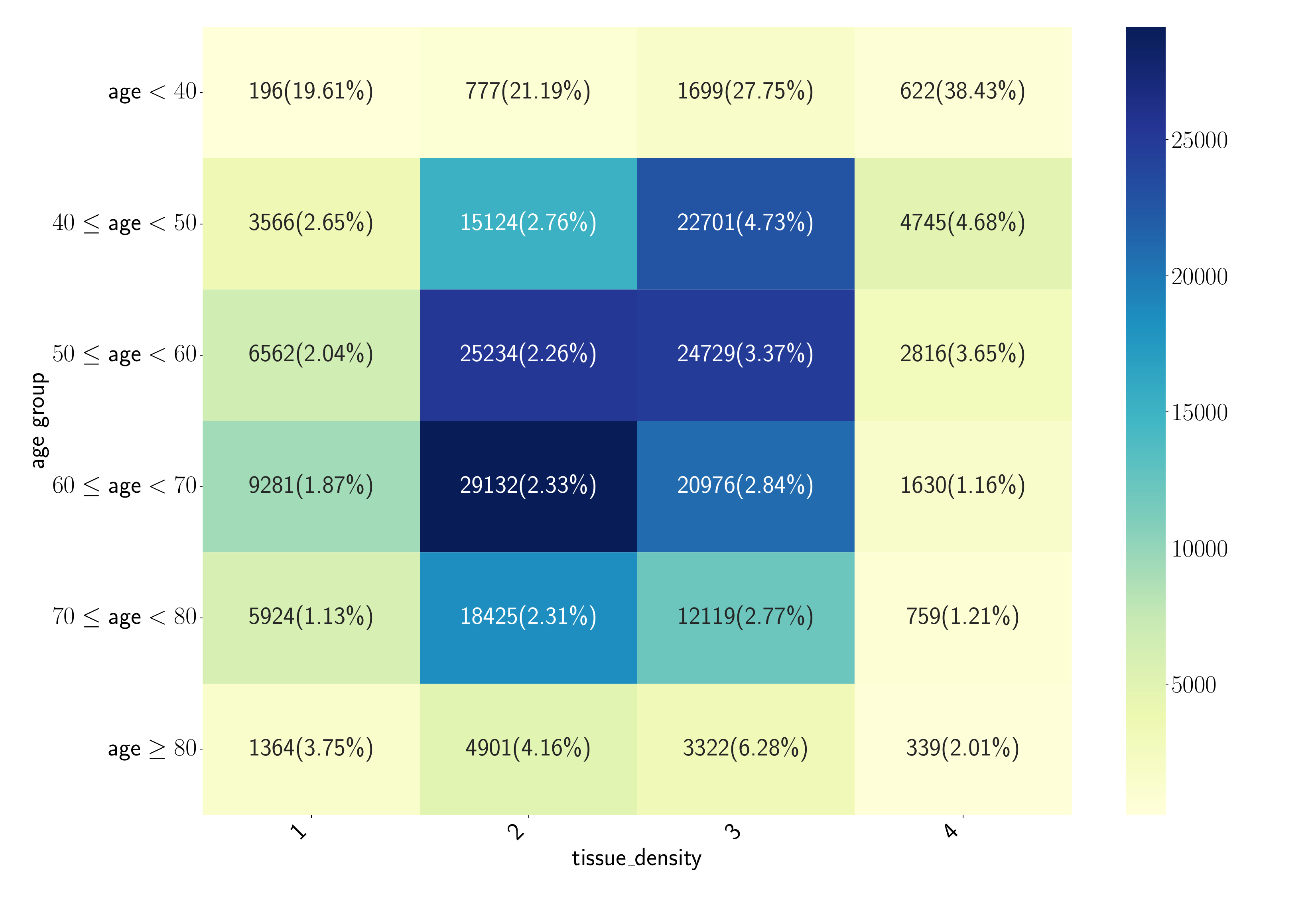}
      \end{subfigure}
    \begin{subfigure}[t]{0.32\textwidth}
      \includegraphics[width=\linewidth]{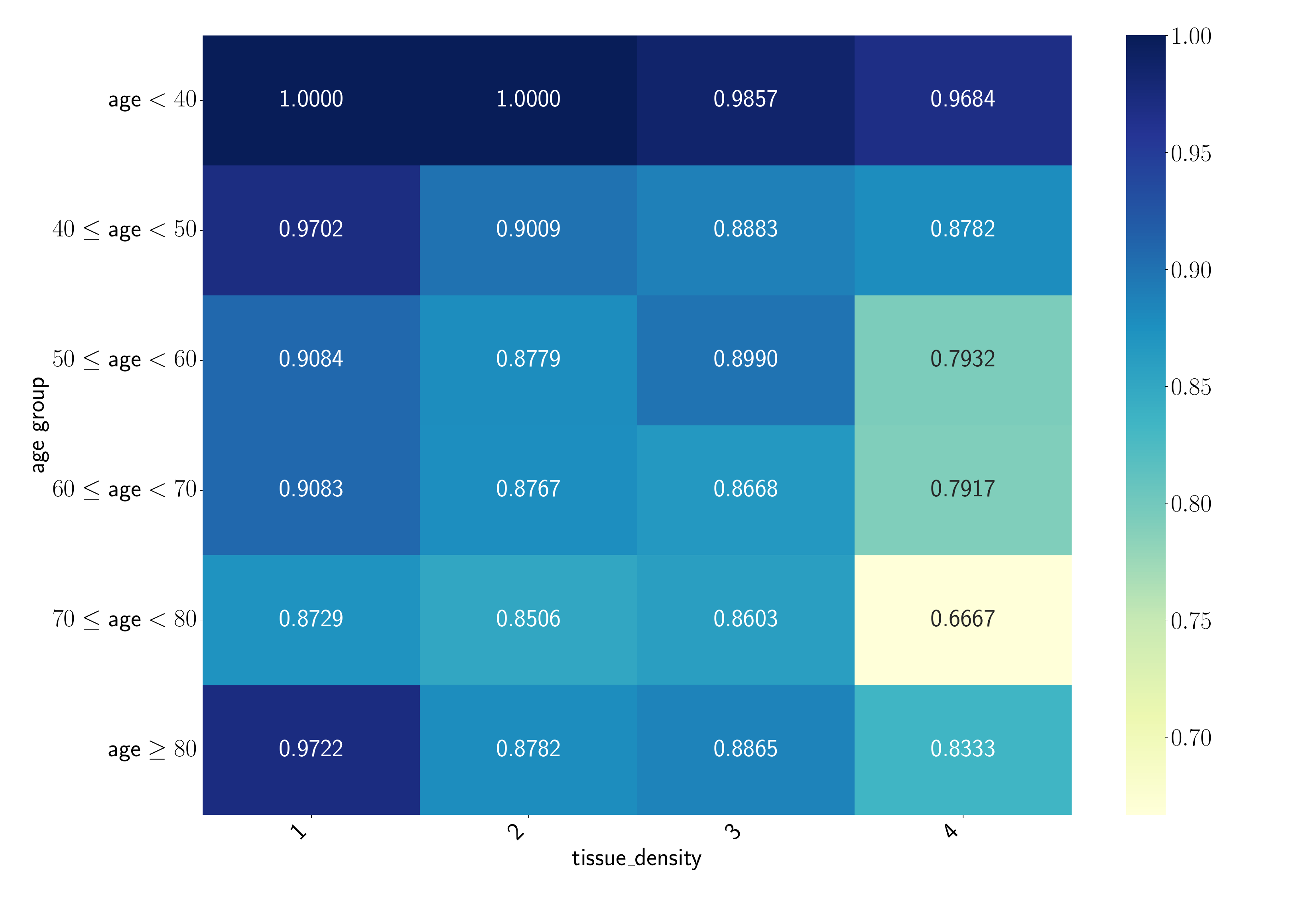}\caption{age group - tissue density}
      \end{subfigure}
    \begin{subfigure}[t]{0.32\textwidth}
      \includegraphics[width=\linewidth]{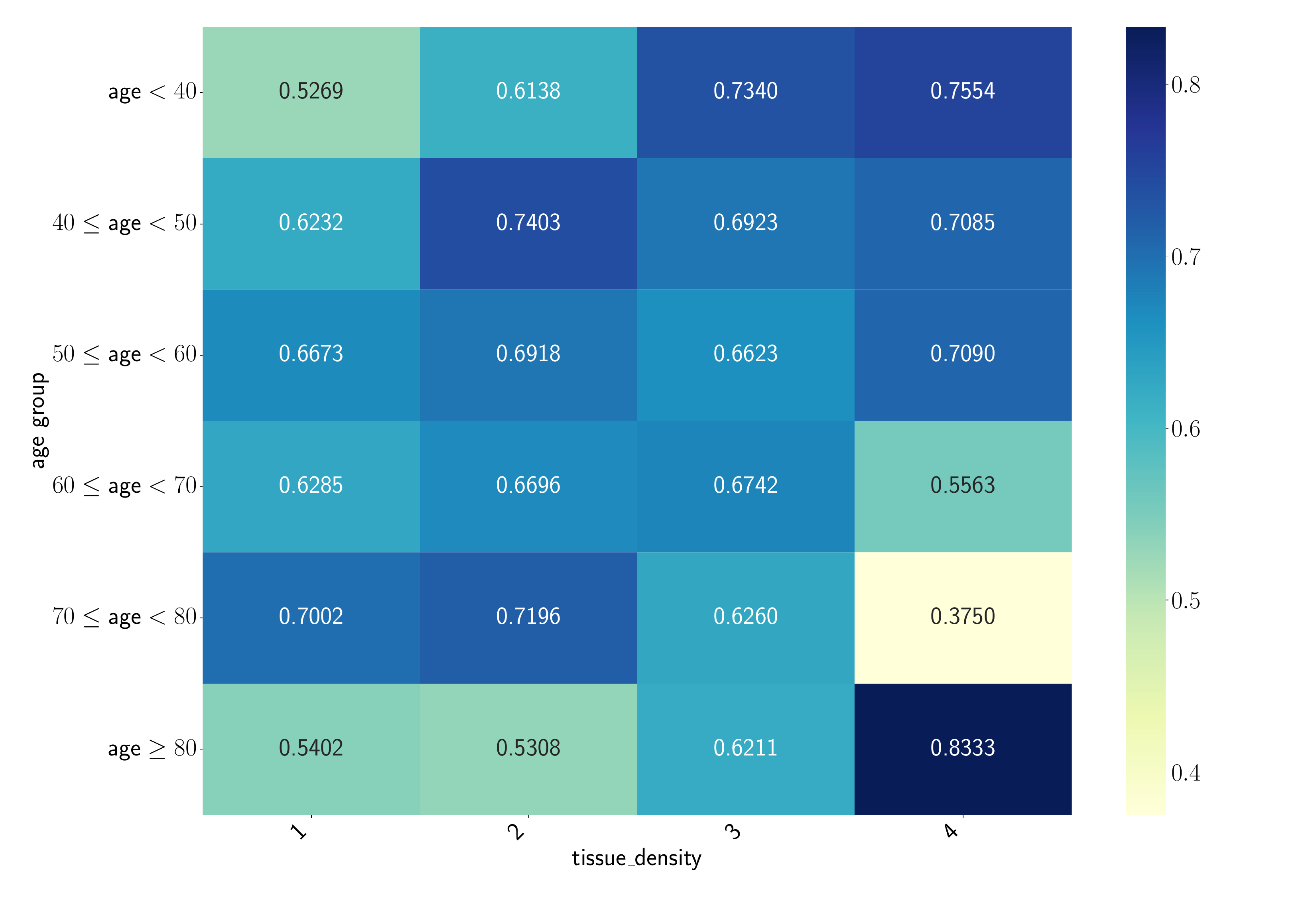}
      \end{subfigure}\\
    \begin{subfigure}[t]{0.32\textwidth}%
      \includegraphics[width=\linewidth]{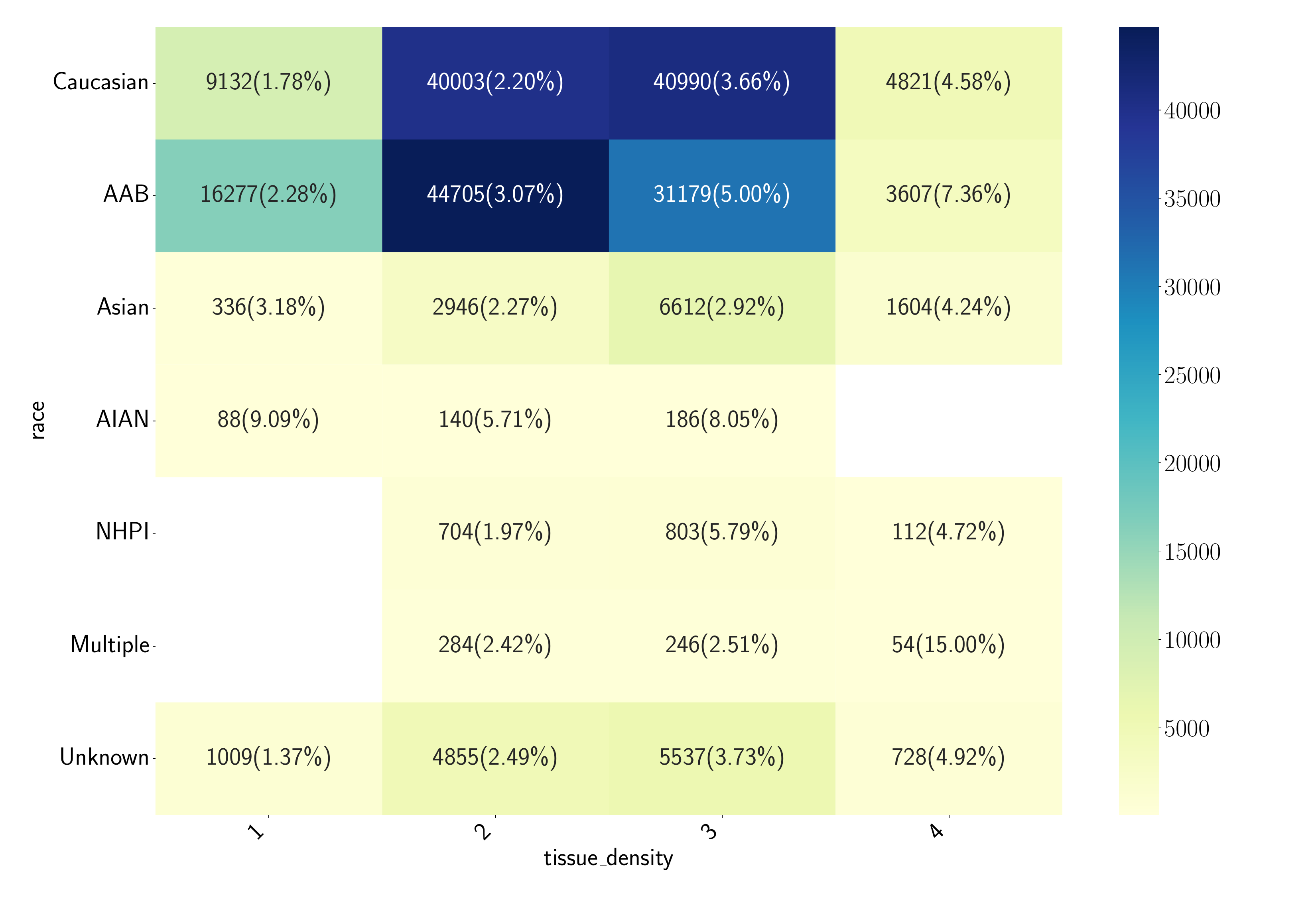}
      \end{subfigure}
    \begin{subfigure}[t]{0.32\textwidth}
      \includegraphics[width=\linewidth]{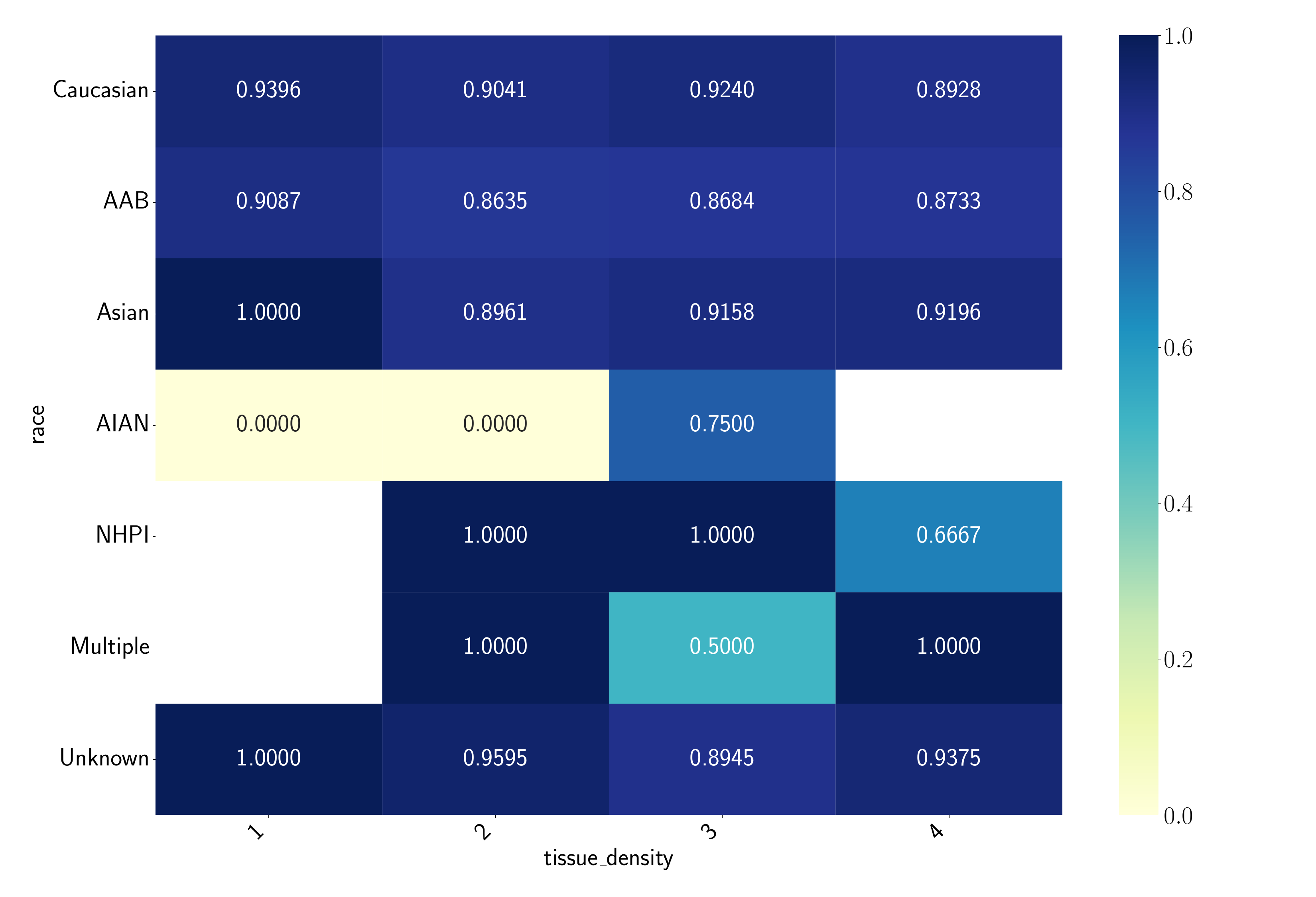}\caption{race-tissue density}
      \end{subfigure}
    \begin{subfigure}[t]{0.32\textwidth}
      \includegraphics[width=\linewidth]{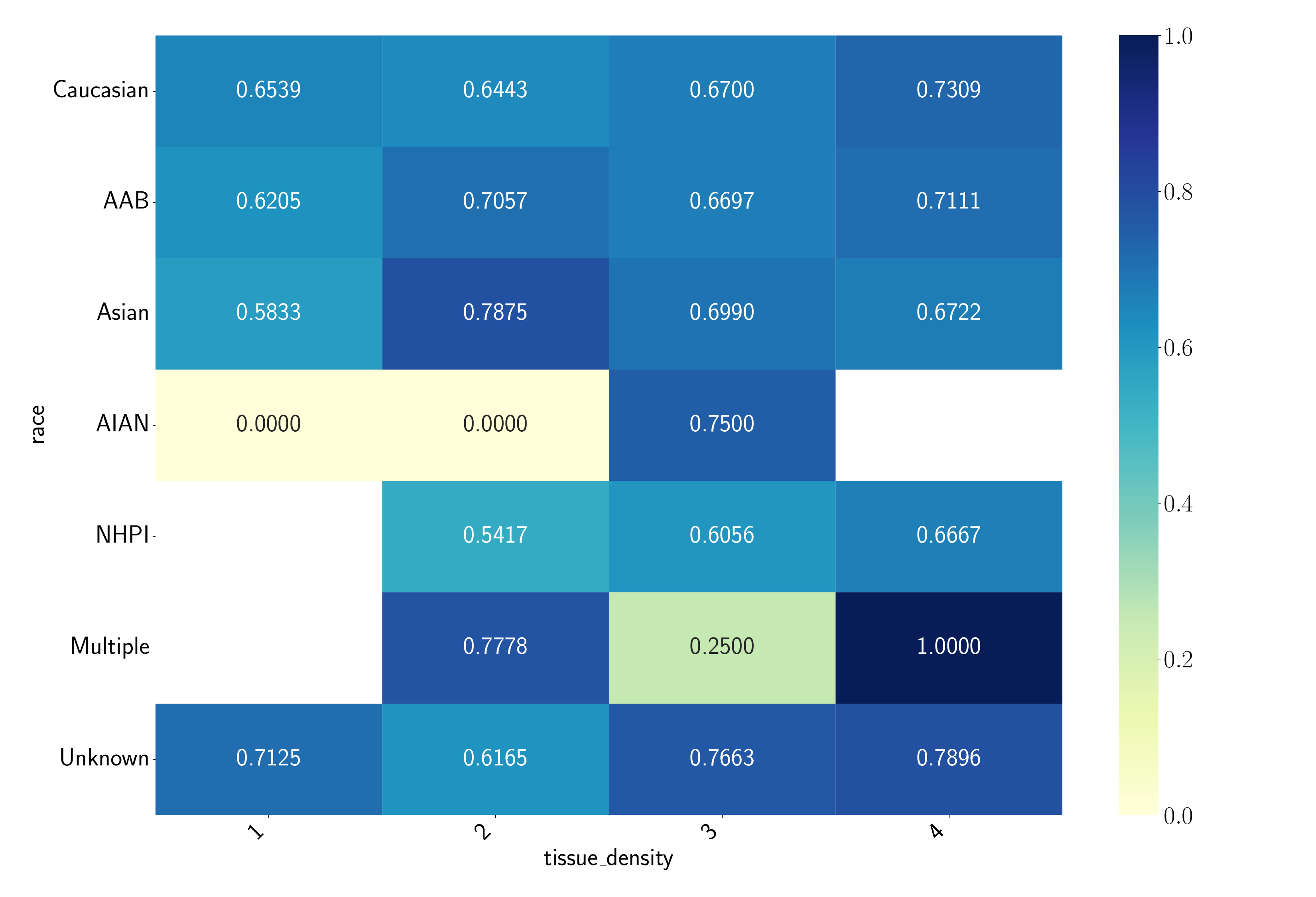}
      \end{subfigure}\\
\caption{The number of samples, PPV, and sensitivity for the joint attribute based subgroups in EMBED, such as (a) age group-race, (b) age group-tissue density and (c) race- tissue density. The blank cells indicate no positive test samples for these groups.}\label{fig:subgroup_age_eth}
\end{figure}

\begin{table*}[hbt]\centering
\caption{Detailed subgroup results on the EMBED dataset}\label{tab:detailed_EMBED}
%\scriptsize
\begin{tabular}{lrrrrrrrr}\hline
Groups &AU- &PPV &sensi- &F1 &speci- &accu- & uncer- \\
 & ROC&&tivity &score &ficity & racy& tainty \\\hline
age $< 40$ &0.9287 &0.9827 &0.7099 &0.8238 &0.9951 &0.9146 &0.0682 \\
$40 \le \text{age} < 50$ &0.9347 &0.8958 &0.7038 &0.7882 &0.9966 &0.9849 &0.0656 \\
$50 \le \text{age} < 60$ &0.9500 &0.8870 &0.6776 &0.7679 &0.9975 &0.9884 &0.0597 \\
$60 \le \text{age} < 70$ &0.9420 &0.8803 &0.6670 &0.7587 &0.9977 &0.9895 &0.06 \\
$70 \le \text{age} < 80$ &0.9513 &0.8573 &0.6773 &0.7566 &0.9973 &0.9898 &0.0657 \\
age $\ge 80$ &0.9482 &0.8904 &0.5741 &0.6975 &0.9964 &0.9761 &0.0763 \\
\hline
Caucasian &0.9484 &0.9173 &0.6676 &0.7726 &0.9982 &0.9884 &0.0618 \\
AAB &0.9407 &0.8747 &0.6831 &0.7670 &0.9961 &0.9841 &0.0643 \\
Asian &0.9395 &0.9136 &0.7038 &0.7940 &0.9979 &0.9890 &0.0639 \\
AIAN &0.9236 &1.0 &0.7083 &0.8167 &1.0 &0.9839 &0.0591 \\
NHPI &0.9259 &1.0 &0.6019 &0.7470 &1.0 &0.9850 &0.0628 \\
Multiple &0.9866 &0.9167 &0.7417 &0.7889 &0.9983 &0.9910 &0.0625 \\
Unknown &0.9552 &0.9207 &0.7139 &0.8030 &0.9980 &0.9889 &0.0605 \\
\hline
density 1 &0.9578 &0.9372 &0.6492 &0.7670 &0.9988 &0.9897 &0.0514 \\
density 2 &0.9463 &0.8829 &0.6798 &0.7680 &0.9976 &0.9892 &0.0585 \\
density 3 &0.9355 &0.8976 &0.6771 &0.7717 &0.9967 &0.9835 &0.0694 \\
density 4 &0.9479 &0.8914 &0.7210 &0.7962 &0.9949 &0.9799 &0.0788 \\
%5 &0.9602 &0.9500 &0.7267 &0.8210 &0.9167 &0.7640 &0.0705\\
\hline
MLO view &0.9393 &0.8881 &0.6428 &0.7455 & 0.9977 &0.9881 &0.0616 \\
CC view &0.9481 &0.9014 &0.7063 &0.7918 &0.9967 &0.9847 &0.0645 \\
\hline
Lorad Selenia & 0.9499 & 0.8346 & 0.7567 & 0.7923 & 0.9964 & 0.9906 & 0.0627\\
Selenia Dimensions & 0.9446 & 0.8985 & 0.6814 & 0.7749 & 0.9972 & 0.9863 & 0.0659\\
S2000D ADS17.4.5 & 0.9530 & 0.9038 & 0.6479 & 0.7534 & 0.9976 & 0.9872 & 0.06\\
S2000D ADS17.5 & 0.9349 & 0.9008 & 0.6181 & 0.7187 & 0.9974 & 0.9861 & 0.0679\\
SEss. ADS53.40 & 0.9413 & 0.8751 & 0.6277 & 0.7222 & 0.9978 & 0.9889 & 0.0625\\
S Pristina & 0.9358 & 0.75 & 0.2604 & 0.3745 & 1.0 & 0.9647 & 0.0656\\
\hline
\end{tabular}
\end{table*}

\begin{table*}[hbt]\centering
\caption{Detailed subgroup results on the RSNA dataset}\label{tab:detailed_RSNA}
%\scriptsize
\begin{tabular}{lrrrrrrrr}\hline
Groups &AUROC &PPV &sensi- &F1 &speci- &accuracy & uncer- \\
 & &&tivity &score &ficity & & tainty \\\hline
age $< 40$ & 0.9994 & 0.8452 & 0.90 & 0.8504 & 0.9979 & 0.9968 & 0.0086\\
$40 \le \text{age} < 50$ &0.9110 & 0.4442 & 0.5111 & 0.4748 & 0.9942 & 0.9898 & 0.0104\\
$50 \le \text{age} < 60$ &0.9330 & 0.54 & 0.5489 & 0.5437 & 0.9928 & 0.9861 & 0.0115\\
$60 \le \text{age} < 70$ &0.9303 & 0.5735 & 0.5126 & 0.5405 & 0.9908 & 0.9797 & 0.0138\\
$70 \le \text{age} < 80$ &0.9076 & 0.7050 & 0.5261 & 0.6026 & 0.9888 & 0.9665 & 0.0199\\
age $\ge 80$ &0.9352 & 0.5750 & 0.5819 & 0.5777 & 0.9770 & 0.9574 & 0.0264\\
\hline
site 1 & 0.9164 & 0.4854 & 0.4684 & 0.4761 & 0.9884 & 0.9767 & 0.0155\\
site 2 & 0.9463 & 0.7294 & 0.6129 & 0.6658 & 0.9954 & 0.9879 & 0.0106\\\hline
density 1 & 0.9075 & 0.4595 & 0.50 & 0.4771 & 0.9895 & 0.9812 & 0.0127\\
density 2 & 0.9101 & 0.4620 & 0.4830 & 0.4714 & 0.9857 & 0.9734 & 0.0169\\
density 3 & 0.9203 & 0.5162 & 0.4504 & 0.4806 & 0.9901 & 0.9778 & 0.0156\\
density 4 & 0.9529 & 0.6062 & 0.42 & 0.4938 & 0.9954 & 0.9860 & 0.0113\\
\hline
MLO view & 0.9312 & 0.5747 & 0.5127 & 0.5412 & 0.9917 & 0.9816 & 0.0140\\           
CC view & 0.9328 & 0.5892 & 0.5499 & 0.5686 & 0.9917 & 0.9823 & 0.0126\\
\hline
machine 21 & 0.9532 & 0.7857 & 0.5909 & 0.6724 & 0.9967 & 0.9891 & 0.0082\\
machine 29 & 0.9377 & 0.7061 & 0.5901 & 0.6413 & 0.9951 & 0.9872 & 0.0119\\
machine 48 & 0.9513 & 0.7143 & 0.6522 & 0.6817 & 0.9945 & 0.9875 & 0.0116\\
machine 49 & 0.9107 & 0.5147 & 0.4646 & 0.4876 & 0.9881 & 0.9744 & 0.0163\\
machine 93 & 0.9637 & 0.2933 & 0.6071 & 0.3953 & 0.9891 & 0.9863 & 0.0117\\
machine 170 & 0.9398 & 0.5042 & 0.6196 & 0.5554 & 0.9844 & 0.9754 & 0.0182\\
machine 190 & 0.8661 & 0.6405 & 0.50 & 0.5304 & 0.9875 & 0.9707 & 0.0180\\
\hline
\end{tabular}
\end{table*}

\subsection{Detailed results based on the attributes}\label{apd:first}
The detailed results on EMBED and RSNA with the range of classification metrics for different subgroups are presented in Table \ref{tab:detailed_EMBED} and Table \ref{tab:detailed_RSNA} respectively. We also present the subgroup analysis based on the joint attributes, such as age-race, race-density and age-density for EMBED in Fig. \ref{fig:subgroup_age_eth}.
%Moreover, the fairness measures evaluated with the pretrained model on EMBED are presented in Table \ref{tab:fairness_pretrained}.

%\subsection{Performance analysis based on the joint attributes}\label{Perf_age-eth}
%In Fig. \ref{fig:subgroup_age_eth}, we present the PPV and sensitivity measures along with the number of samples and prevalence (\%) for the joint attribute based subgroups. The findings are mostly similar to the findings observed in Fig. \ref{fig:subgroup_pop}. For example, PPV for the AAB population groups with age $\ge 40$ are lower than most Caucasian and Asian counterparts. We observe a drop in sensitivity for the `age $\ge 80$' group than other groups for the Caucasian population. There are only a few number of positive samples for the AIAN, NHPI and Multiple groups with various age and for the Asian population with age $>70$. We can observe a high PPV and low sensitivity for most NHPI and Multiple subgroups.

\subsection{Performance Monitoring}\label{sec:app_perf_moni}
The tuned $k$ value, the number of false positives, and detection delays are presented in Table \ref{tab:moni_results}. We can observe that there are only a few false positives with only a minimal average detection delay.

%\begin{table}[t]\centering
%\caption{Fairness evaluated with the \cite{dangnh0611} model on the EMBED}\label{tab:fairness_pretrained}
%\scriptsize
%\begin{tabular}{lp{1.8cm}p{1.4cm}}\hline
%attribute &  Equal opportunity $\downarrow$ &Predictive parity $\uparrow$ \\\hline
%age group &\textbf{0.2540} &\textbf{0.2463} \\
%race &\textbf{0.3739} &\textbf{0.5901} \\
%tissue density &\textbf{0.1799} &\textbf{0.6270} \\
%view position &0.1060 &\textbf{0.4762} \\
%\hline
%\end{tabular}
%\end{table}

\begin{table}[t]\centering
\caption{Sensitivity monitoring results with design choices}\label{tab:moni_results}
%\scriptsize
\begin{tabular}{p{3cm}p{0.8cm}p{1.4cm}p{1.2cm}}\hline
under-performing group & Tuned $k$ & FAR & Detection Delay \\\hline
NHPI & 0.06 & 1 &1\\
`age $\ge 80$' & 0.06 & 4 &2\\
%density of 1.0 & 1.25 & 1 & 0 \\
%MLO view & 1.25  &1 & 3 \\
\hline
\end{tabular}
\end{table}

%\begin{figure}[t]
%\centering
%    \begin{subfigure}[t]{0.7\textwidth}
%      \includegraphics[width=\linewidth]{subgroup_figures/updated_cusum_embed_recall_age_group_0.06.pdf}\caption{}\label{fig:moni_age}
%      \end{subfigure}
    %\begin{subfigure}[t]{0.7\textwidth}
    %  \includegraphics[width=\linewidth]{subgroup_figures/CUSUM_fold2_metric_recall_atttissue_density_up_1.25.pdf}\caption{}\label{fig:moni_tisden}
    %  \end{subfigure}\\
    %\begin{subfigure}[t]{0.7\textwidth}
    %  \includegraphics[width=\linewidth]{subgroup_figures/CUSUM_fold2_metric_recall_attviewposition_up_1.25.pdf}\caption{}\label{fig:moni_view}
    %  \end{subfigure}

%\caption{CUSUM based sensitivity monitoring charts for a tuned $k$. Here, we consider the `age $\ge 80$' group as the underperforming group in batch formation.}\label{fig:perf_moni2}
%\end{figure}

\end{appendices}


\end{document}